\documentclass[letterpaper]{article} 
\usepackage{aaai2026}  
\usepackage{times}  
\usepackage{helvet}  
\usepackage{courier}  
\usepackage[hyphens]{url}  
\usepackage{graphicx} 
\usepackage{natbib}  
\usepackage{caption} 
\usepackage{algorithm}
\usepackage{algorithmic}
\usepackage{booktabs}
\usepackage{xcolor}
\usepackage{verbatim}
\usepackage{newfloat}
\usepackage{pifont}
\usepackage{graphicx}
\usepackage{amsmath}
\usepackage{amssymb}
\usepackage{booktabs}
\usepackage{caption}
\usepackage{subcaption}
\usepackage{xcolor}
\usepackage{subcaption}
\usepackage{fancyvrb}
\usepackage{fvextra}
\fvset{breaklines=true}
\usepackage{listings}
\usepackage{enumitem}
\usepackage{multirow}
\usepackage{arydshln}
\newcommand{\cdashlinelr}[1]{%
  \noalign{\vskip 2pt}   
  \cdashline{#1}[.4pt/2pt] 
  \noalign{\vskip 2pt}   
}
\setlist[itemize,enumerate]{leftmargin=*}
\DeclareCaptionStyle{ruled}{labelfont=normalfont,labelsep=colon,strut=off} 
\floatstyle{ruled}
\newfloat{listing}{tb}{lst}{}
\floatname{listing}{Listing}
\title{SheetBrain: A Neuro-Symbolic Agent for Accurate Reasoning over~Complex~and~Large~Spreadsheets}

\author{
    Ziwei Wang\textsuperscript{\rm 1}\thanks{Work done during internship at Microsoft.}\thanks{Ziwei and Jiayuan contributed equally to this work.},
    Jiayuan Su\textsuperscript{\rm 2*†},
    Mengyu Zhou\textsuperscript{\rm 5}\thanks{Corresponding author (mezho@microsoft.com).},
    Huaxing Zeng\textsuperscript{\rm 3*},
    Mengni Jia\textsuperscript{\rm 4*},\\
    Xiao Lv\textsuperscript{\rm 5},
    Haoyu Dong\textsuperscript{\rm 5},
    Xiaojun Ma\textsuperscript{\rm 5},
    Shi Han\textsuperscript{\rm 5},
    Dongmei Zhang\textsuperscript{\rm 5}
}

\begin{document}
\nocopyright
\affiliations {
    \textsuperscript{\rm 1}Carnegie Mellon University
    \textsuperscript{\rm 2}Zhejiang University
    \textsuperscript{\rm 3}Brown University\\
    \textsuperscript{\rm 4}University of Cambridge
    \textsuperscript{\rm 5}Microsoft Research
}



\maketitle

\begin{abstract}
Understanding and reasoning over complex spreadsheets remain fundamental challenges for large language models (LLMs), which often struggle with accurately capturing the complex structure of tables and ensuring reasoning correctness. In this work, we propose SheetBrain, a neuro-symbolic dual-workflow agent framework designed for accurate reasoning over tabular data, supporting both spreadsheet question answering and manipulation tasks. SheetBrain comprises three core modules: an understanding module, which produces a comprehensive overview of the spreadsheet—including sheet summary and query-based problem insight to guide reasoning; an execution module, which integrates a Python sandbox with preloaded table-processing libraries and an Excel helper toolkit for effective multi-turn reasoning; and a validation module, which verifies the correctness of reasoning and answers, triggering re-execution when necessary. We evaluate SheetBrain on multiple public tabular QA and manipulation benchmarks, and introduce SheetBench, a new benchmark targeting large, multi-table, and structurally complex spreadsheets. Experimental results show that SheetBrain significantly improves accuracy on both existing benchmarks and the more challenging scenarios presented in SheetBench. Our code is publicly available at https://github.com/microsoft/SheetBrain.
\end{abstract}

\section{Introduction}
Large language models (LLMs) have achieved remarkable success across diverse natural language understanding and generation tasks, particularly in domains involving unstructured text~\cite{xu2025chain, wang2024q, pan2024chain}. However, their ability to understand and reason over complex spreadsheet data for question answering (QA) and manipulation remains a fundamental challenge~\cite{zhu2025sheetmind, chen2024sheetagent, li2023sheetcopilot}. Real-world spreadsheets—especially Excel-like files—often contain multi-table layouts, hierarchical structures, lengthy content, and other complex formatting elements~\cite{wu2025realhitbench, li2024mimotable, zhao2022multihiertt, cheng2021hitab}. These factors significantly increase the difficulty of accurately capturing the table semantics and performing reliable reasoning, limiting the effectiveness of current LLM-based approaches.

\begin{table}[!htb]
\centering
\resizebox{\columnwidth}{!}{%
\begin{tabular}{lccccc}
\toprule
\textbf{Model} & \textbf{Complex} & \textbf{Multi-table} & \textbf{Large Sheet} & \textbf{Edit} & \textbf{Total} \\
\midrule
\textbf{Test Cases} & 21 & 20 & 20 & 8 & 69 \\

\multicolumn{6}{l}{\textcolor{gray}
{\textit{Vanilla LLM:}}} \\
gpt-4.1     & 16 & 17 & 1  & -- & 34 \\
\cdashlinelr{1-6}
\multicolumn{6}{l}{\textcolor{gray}{\textit{Proprietary Agents:}}} \\
BizChat Analyst       & 18 & 14 & 9  & 6  & 49 \\
ChatGPT (4o)          & 13 & 8  & 9  & -- & 34 \\
\cdashlinelr{1-6}
{\textcolor{gray}{\textit{Open-Sourced Agents:}}} \\
StructGPT (4.1)       & 12 & 1 & 0 & -- & 13 \\
SheetAgent (4.1)      & 11 & 10 & 10 & 4 & 35 \\
\textbf{SheetBrain (4.1, ours)} & \textbf{20} & \textbf{18} & \textbf{11} & \textbf{6} & \textbf{55} \\
\bottomrule
\end{tabular}
}
\caption{Comparison results on our proposed benchmark \textit{Sheetbench}. Reasoning over complex, multi-table, and large spreadsheets remains a significant challenge for existing LLM-based approaches.}
\label{tab:model-comparison0}
\vspace{-4mm}
\end{table}

Prior works~\cite{chen2024sheetagent, jiang2023structgpt, li2023sheetcopilot} have shown that LLMs can perform QA and manipulation effectively on simple tables and handle basic instructions. However, they often struggle with challenging scenarios involving complex and large spreadsheets. This is empirically demonstrated in the examples summarized in Table~\ref{tab:model-comparison0}, where comparison results confirm that LLM-based methods still have significant room for improvement. From these cases, we identify three key findings: (1) These methods do not explicitly analyze the spreadsheet structure or understand the query context beforehand, so their reasoning tends to be blind and inefficient; (2) The proprietary agent BizChat~\cite{microsoft365copilot} shows better performance on complex and large tables compared to vanilla LLMs by integrating symbolic code execution sandboxes; (3) The agents often rely on multi-step reasoning that is prone to getting stuck in local or narrow perspectives, lacking mechanisms for reflection.

\begin{figure*}[t] 
    \centering 
    \includegraphics[width=1\textwidth]{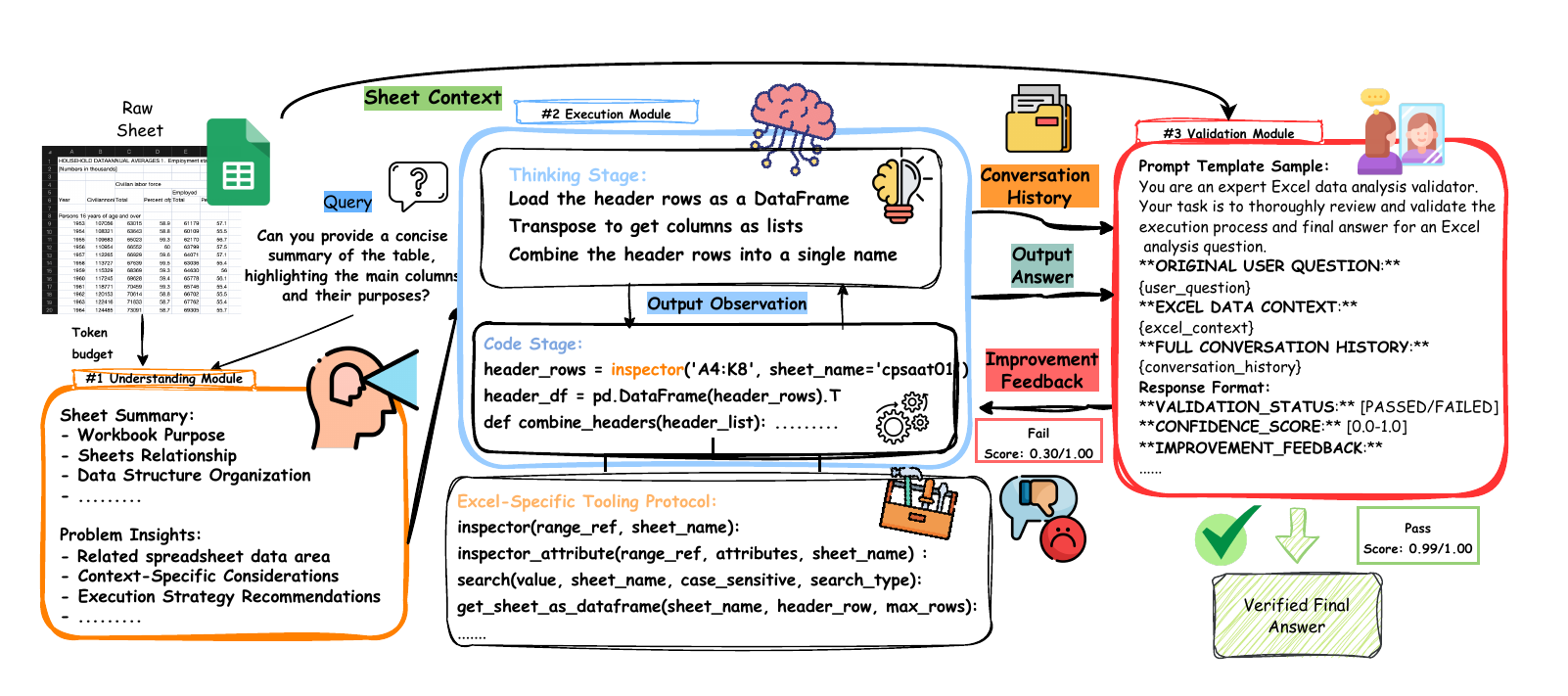} 
    \caption{SheetBrain Pipeline. SheetBrain adopts a robust three-stage pipeline: understand, execute, and validate. It first constructs a global, query-aware understanding of the spreadsheet, then performs tool-augmented reasoning in a Python sandbox, and finally validates and refines its output through iterative self-correction.} 
    \label{fig:main_figure} 
\end{figure*}


Inspired by these findings, we present SheetBrain, a unified neuro-symbolic framework that enhances LLMs’ ability to understand and reason over complex spreadsheets for QA and manipulation. As shown in Figure~\ref{fig:main_figure}, SheetBrain begins with an understanding module that generates a comprehensive overview of the spreadsheet, including a sheet summary and query-specific problem insights, which serve as a scaffold for high-level reasoning. Next, an execution module leverages a specialized Python sandbox integrated with preloaded table-processing libraries and a custom Excel helper toolkit to enable tool-augmented, multi-turn reasoning of tabular data. Finally, a validation module verifies the correctness of both reasoning and output, triggering re-execution with improvement feedback when necessary to avoid getting stuck in local errors and to improve reliability. These components bridge the gap between challenging tabular cases and reasoning capabilities of LLM agents, enabling more robust, accurate, and interpretable performance across diverse spreadsheet scenarios.

To evaluate the effectiveness of SheetBrain, we conduct extensive experiments on several public spreadsheet QA and manipulation datasets. Additionally, we introduce SheetBench, a new benchmark specifically curated to reflect challenges such as large sheets, multi-table layouts, and deeply nested structures. Experimental results show that SheetBrain not only outperforms strong baselines—including vanilla LLMs and spreadsheet LLM-based agents—across public benchmarks, but also generalizes significantly better to the complex scenarios presented in SheetBench. 

We summarize our contribution as follows:

\begin{itemize}[topsep=1pt, before=\setlength{\itemsep}{1pt}, parsep=1pt, leftmargin=*]
\item We propose \textbf{SheetBrain}, a novel spreadsheet agent based on an understand-execute-validate pipeline. It first generates a comprehensive overview of the spreadsheet, then passes it to a hybrid neuro-symbolic workflow execution module that integrates a Python sandbox along with dedicated libraries and toolkits for reasoning, and finally to a validation module for verification, triggering re-execution with improvement feedback when necessary.

\item We introduce \textbf{SheetBench}, a new benchmark comprising 69 challenging real-world spreadsheet QA and manipulation tasks, specifically designed to evaluate LLM performance on large-scale, structurally complex spreadsheets featuring multi-table layouts. Experimental results show that SheetBrain achieves state-of-the-art performance, outperforming vanilla LLMs such as GPT-4.1~\cite{openai-gpt4-1}, o4-mini~\cite{openai-o3-o4-mini}, DeepSeek-R1~\cite{guo2025deepseek}, and Qwen-3-32B~\cite{yang2025qwen3}, as well as prior open-source and proprietary spreadsheet agents across three public benchmarks. On SheetBench, SheetBrain demonstrates significantly superior performance, highlighting its capability to handle challenging sheets.

\item We conduct extensive experiments and analyses, including module ablations that demonstrate the importance of both understanding and validation, evaluations of different tool roles highlighting the neuro-symbolic workflow, and investigations into effective spreadsheet encoding strategies. These provide valuable insights and guidance for future spreadsheet agent designs. Additionally, we present case studies comparing existing proprietary agents with SheetBrain, revealing key findings.

\end{itemize}

\section*{SheetBrain}
SheetBrain employs a robust three-stage pipeline: understand, execute, and validate. In the understanding stage, the LLM generates both a sheet summary and task-specific information to develop a global comprehension of the spreadsheet, laying the foundation for the subsequent stages. During execution, the LLM operates within a Python sandbox environment equipped with preloaded table-processing libraries and a customized Excel helper toolkit, enabling effective hybrid neuro-symbolic reasoning. In the validation stage, the LLM evaluates whether the execution reasoning and results are reasonable. If inconsistencies arise, the execution module incorporates feedback from the validation module to refine its reasoning and re-execute, iteratively enhancing the outcome.

\subsection{Understanding Module}
\begin{figure}[t] 
    \centering 
    \includegraphics[width=0.5\textwidth]{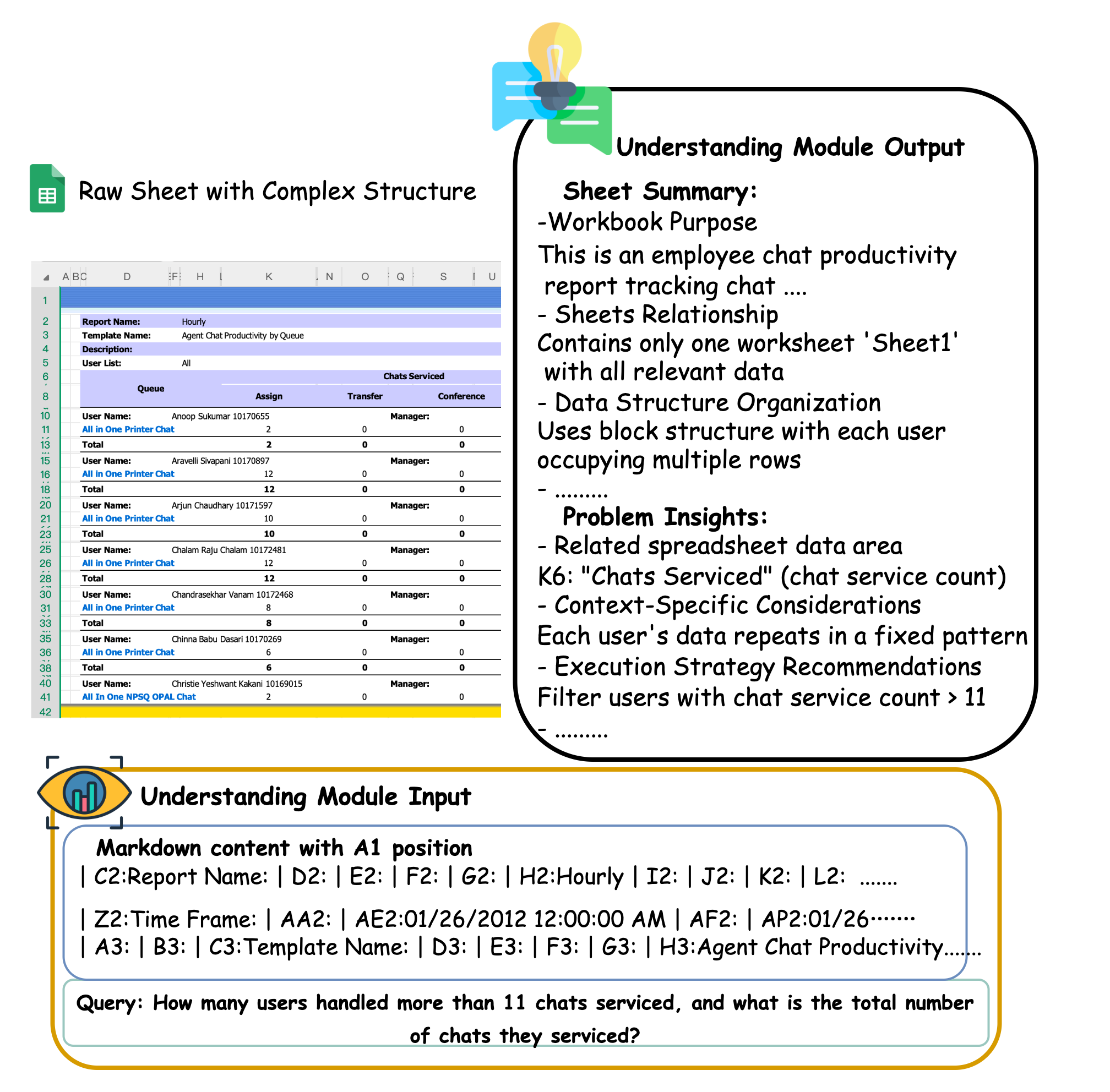} 
    \caption{The workflow of the understanding module.} 
    \label{fig:understanding_figure} 
\end{figure}

The understanding module serves as SheetBrain’s cognitive front-end, transforming a user query and raw spreadsheet data into valuable prior knowledge. As shown in Figure~\ref{fig:understanding_figure}, given the raw spreadsheet data and user query, we prompt the LLM to generate a sheet summary that includes the sheet’s purpose, object relationships and structures, as well as problem insights indicating which parts of the spreadsheet data are relevant to the query. To balance providing sufficient context with computational constraints—especially for ultra-large spreadsheets—it employs dynamic token budget management, allocating a token budget per sheet to capture essential information while avoiding context window overflow. Within this budget, we utilize an enhanced markdown serialization that encodes cell content, cell position notation, and merged cell annotations. This approach preserves crucial 2D spatial context, enabling the agent to perceive the sheet’s layout in a manner that mirrors human visual interpretation and reasoning. The detailed information generated by this module is listed as follows:
\begin{itemize}
    \item \textbf{Sheet Summary:} An overview outlining the workbook’s overall objective, key relationships (e.g., between sheets), and data structure organization, as well as conducting an in-depth structural analysis, for example, as shown in Figure~\ref{fig:understanding_figure}, by recognizing that “each user’s data occupies a distinct multi-row block”, the summary guides the execution module to iterate over blocks rather than individual rows.

    \item \textbf{Problem-Specific Insights:} Insights derived directly from the user's query, which include identifying important rows and columns, recommending focus areas, and suggesting strategic approaches for the execution module. The recommendation to \textit{``Filter users with chat service count $> 11$''} exemplifies this by providing a direct, executable strategy that prunes the search space, ensuring the final execution is both efficient and accurate.
\end{itemize}

\subsection{Execution Module}

The execution module serves as the operational core of SheetBrain, where task decomposition, code generation, and data manipulation occur. The module is built upon three main components:  an Excel-specific tooling protocol; a dual neuro-symbolic dataflow architecture equipped with a Python sandbox; and an iterative reasoning and execution cycle. This design allows the agent to have one or several interactive turns to perform neuro-based reasoning or treat Excel files as manipulable variables, providing a persistent and controlled environment for symbolic computation.

\paragraph{Excel-Specific Tooling Protocol}  
Our Excel-specific tooling protocol is empirically designed based on systematic analysis of common operational patterns and failure modes encountered by agents working with spreadsheets. The protocol encapsulates these insights into a set of specialized Python functions that integrate neural reasoning with symbolic execution. It enables seamless conversion of spreadsheet data into standard Python data structures, facilitating complex analyses through symbolic code. For example, the protocol incorporates robust search capabilities that support exact matches as well as partial and whitespace-tolerant queries, enhancing resilience to noisy or ambiguous inputs. Additionally, it provides fine-grained cell inspection, including extraction of visible values, formatting attributes (e.g., cell color, font), and underlying formulas. These features allow agents to leverage critical visual and structural cues essential for understanding spreadsheets in real-world scenarios.

\paragraph{Symbolic Dataflow Architecture}  
Unlike conventional agents that rely on neural dataflow—embedding raw tool outputs back into the LLM’s context—SheetBrain employs a symbolic dataflow architecture. This design handles queries by storing the sheet in a named variable, such as a pandas DataFrame, and generating code that directly references this variable to perform operations like counting the occurrences of values in a specific column, enabling accurate and precise answers (see Appendix Case 2). In contrast, neural dataflow approaches return thousands of raw data entries into the LLM’s context, consuming excessive tokens that may exceed the context limit and hinder accurate computation. By offloading operations to the Python interpreter, SheetBrain ensures computational fidelity and scalable memory, allowing it to process large spreadsheets where neural dataflow agents face fundamental limitations.

\paragraph{Iterative Reasoning and Execution Cycle}  
The execution module does not follow a single-shot execution paradigm. Instead, it adopts an iterative cycle alternating between reasoning and execution. In each iteration, the agent performs high-level reasoning based on the user query, the current symbolic state, and prior outputs, then generates corresponding Python code. This code is executed within a sandbox environment, which updates the symbolic state with new or modified variables. The resulting outputs—whether values, transformed DataFrames, error messages, or confirmations—are returned as feedback to the agent. This feedback guides the agent’s subsequent reasoning, enabling it to refine its strategy, correct errors, or advance to the next logical subtask. For example, the agent may first load a sheet, then apply a filter, and subsequently aggregate the filtered data. This cycle continues until the agent determines that the task is complete and the current output satisfies the user query.

\subsection{Validation Module}
The validation module serves as the final and critical stage in the SheetBrain pipeline, functioning as a quality assurance and self-correction mechanism. It is invoked after the execution module generates a provisional answer, and is responsible for systematically evaluating the result before producing the final output. This evaluation is guided by a structured checklist and is informed by multiple inputs: the spreadsheet preview, the reasoning trace, the execution logs, the original user query, and the proposed answer. These inputs are cross-referenced to verify logical consistency, correctness, and relevance. The checklist assesses key dimensions such as data handling (e.g., extraction, transformation accuracy) and answer quality (e.g., completeness, format adherence, and alignment with the user query). The module outputs a binary decision—pass or fail—along with a confidence score ranging from 0 to 1. In case of failure, it provides detailed diagnostic feedback and concrete suggestions for improvement, which are forwarded to the execution module. This feedback initiates another execution cycle, enabling the agent to revise its approach and correct errors. The process repeats iteratively until the validation module returns a positive judgment. Only then is the final answer delivered to the user. This closed-loop feedback architecture is crucial for achieving high reliability and precision in complex spreadsheet reasoning tasks.

\section{SheetBench}

\begin{figure}[t] 
    \centering 
    \includegraphics[width=0.4\textwidth]{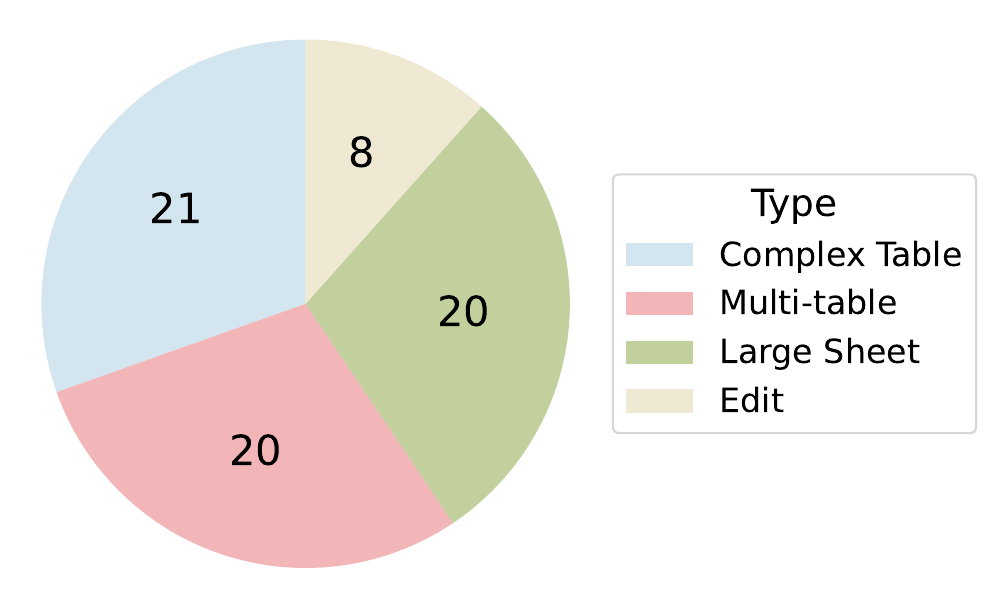} 
    \caption{Statistics of SheetBench. The numbers shown in the pie chart represent the counts for each data type.} 
    \label{fig:sheetbench_chart} 
\end{figure}

To address the gap in existing spreadsheet benchmarks, which focus on small, well-structured tables and overlook real-world complexity, we carefully developed SheetBench—a comprehensive benchmark designed to evaluate methods’ capabilities in challenging spreadsheet QA and manipulation tasks. Compiled from 11 public sources—including HiTab~\cite{cheng2021hitab}, MimoTable~\cite{li2024mimotable}, RealHiTBench~\cite{wu2025realhitbench}, MultiHiertt~\cite{zhao2022multihiertt}, DabStep~\cite{egg2025dabstep}, SheetCoPilot~\cite{li2023sheetcopilot}, SpreadsheetBench~\cite{ma2024spreadsheetbenchchallengingrealworld}, SheetAgent~\cite{chen2024sheetagent}, NCSE~\cite{ncse2025}, MMQA~\cite{wu2025mmqa}, and CodaBench~\cite{xu2022codabench}—it contains 69 cases carefully selected by human annotators for their challenge to existing LLM agents such as ChatGPT~\cite{openai2023chatgpt} and BizChat~\cite{microsoft365copilot}. During the curation process, we also corrected queries and sheet issues that could cause misleading errors, ensuring these 69 cases maintain high quality and reliability. The benchmark systematically assesses methods across four key challenges—understanding complex structures, reasoning over multiple tables, scaling to large sheets, and performing spreadsheet editing. Covering a wide range of tasks from QA to manipulation involving formula propagation and layout modifications, SheetBench provides a rigorous and holistic testbed for advancing spreadsheet intelligence. Figure~\ref{fig:sheetbench_chart} summarizes the benchmark’s statistics.

        

\begin{table*}[t]
\centering
\resizebox{\textwidth}{!}{
\begin{tabular}{lccccccccc}
\toprule
\textbf{Model} & \multicolumn{1}{c}{\textbf{multihiertt}} & \multicolumn{3}{c}{\textbf{spreadsheetbench}} & \multicolumn{4}{c}{\textbf{realhitbench}} & \multicolumn{1}{c}{\textbf{sheetbench}} \\
\cmidrule(lr){2-2} \cmidrule(lr){3-5} \cmidrule(lr){6-9} \cmidrule(lr){10-10}
 &  & cell-level & sheet-level & \textbf{Overall} & Fact Checking & Data Analysis & Numerical Reasoning & \textbf{Overall} &  \\
\midrule
\multicolumn{10}{l}{\textit{Vanilla Models}} \\
gpt-4.1 & 53.5 & NA & NA & NA& 75.0 & 71.1 & 70.4 & 70.0 & 50.7 \\
o4-mini & 54.2 & NA & NA & NA& 76.7 & 71.4 & 77.0 & 71.7 & 52.0 \\
4o & 51.8 & NA & NA &NA & 74.3 & 63.9 & 71.4 & 67.7 & 50.7 \\
Qwen3-32b & 50.2 & NA & NA & NA& 73.9 & 71.4 & 66.7 & 68.1 & 50.7 \\
Deepseek-R1 & 51.1 & NA & NA &NA & 69.6 & 50.0 & 66.7 & 64.6 & 52.0 \\
\midrule
\multicolumn{10}{l}{\textit{Other Models}} \\
StructGPT & 13.0 & NA & NA & NA & 44.3 & 30.8 & 42.9 & 40.3 & 21.3\\
SheetAgent & 35.3 & 14.3 & 33.7 & 21.8 & 70.1 & 31.8 & 68.6 & 58.8 & 50.8\\
\textbf{SheetBrain (ours)} & 62.6 & 35.4 & 37.8 & 36.4 & 85.5 & 71.8 & 73.6 & 78.3 & 80.3\\
\bottomrule
\label{Main Results}
\end{tabular}
}
\caption{Main Results: Performance of various models across four evaluation datasets. Each dataset may have multiple sub-tasks (e.g., cell/sheet, fact-checking, etc.). The evaluation for sheetbench focuses solely on question-answering tasks, and the visualization category has been excluded from realhitbench.}
\label{tab:hierarchical-benchmark}
\end{table*}


\section{Experiments and Analysis}
\subsubsection{Experimental Setup}
\paragraph{Benchmarks} To comprehensively evaluate the performance of SheetBrain, we conducted extensive experiments across multiple benchmarks. Our evaluation covers three public datasets: MultiHiertt~\cite{zhao2022multihiertt} (for numerical reasoning over multi-hierarchical tabular and textual data), RealHiTBench~\cite{wu2025realhitbench} (featuring complex, real-world hierarchical tables for LLM-based table analysis), and SpreadsheetBench~\cite{ma2024spreadsheetbenchchallengingrealworld} (a challenging benchmark for real-world spreadsheet manipulation tasks), all of which allowed us to evaluate capabilities in table-based reasoning and operations. Additionally, we introduce our own benchmark, SheetBench, specifically designed to address the challenges posed by large, multi-table, and structurally complex spreadsheets. 

\paragraph{Baselines} Our comparison targets include two categories of models: (1) vanilla LLMs, encompassing several open-source and closed-source models accessed via API, including GPT-4.1~\cite{openai-gpt4-1}, o4-mini~\cite{openai-o3-o4-mini}, 4o~\cite{hurst2024gpt}, Qwen-3-32b~\cite{yang2025qwen3}, and DeepSeek-R1~\cite{guo2025deepseek}; and (2) specialized spreadsheet agents, namely StructGPT and SheetAgent. To ensure a fair comparison, all agent-based models, including our SheetBrain, utilize GPT-4.1 as their backbone LLM. 
In our implementation of SheetBrain, we set a 10,000-token budget for the initial data preview. The spreadsheet's structure is serialized using a Markdown format that includes A1-style cell position notation and explicit indicators for merged cells. 

\paragraph{Evaluator} For SpreadsheetBench, we use its official evaluators to assign scores. For evaluating MultiHierTT, RealHiTBench, and SheetBench, we employ an automated LLM-as-judge evaluator~\cite{li2024llms}, co-developed alongside the SheetBench dataset, to ensure consistent and scalable correctness assessment. Our evaluator prompt is provided in the Appendix.

\subsection{Main Results}
As shown in Table~\ref{tab:hierarchical-benchmark}, across all benchmarks, SheetBrain achieves substantial performance gains over existing methods. It attains state-of-the-art results on all QA and manipulation benchmarks, and also demonstrates superior performance on the more challenging cases in SheetBench.
\paragraph{QA} Compared with vanilla LLMs, GPT-4.1-based SheetBrain improves accuracy by 9.1\% over vanilla GPT-4.1, and also outperforms other LLMs and agents in MultiHiertt. In RealHitBench, SheetBrain improves overall performance by 8.3\%, with a notable gain of over 10\% on fact-checking, as well as improvements in data analysis and numerical reasoning, compared to vanilla GPT-4.1. This suggests that our approach is particularly effective in scenarios that demand precise cross-referencing and consistency checks, which may benefit from the inherent validation mechanisms in our design. The relatively smaller gains in data analysis and numerical reasoning imply that these tasks are more execution-driven and less dependent on iterative validation.
\paragraph{Manipulation} Within the set of baselines, only SheetAgent possesses table-editing capabilities; accordingly, we conduct a direct comparison between SheetBrain and SheetAgent. Experimental results demonstrate that SheetBrain surpasses SheetAgent by nearly 15\% in accuracy, with even greater gains observed at the cell level. These findings underscore the superior fine-grained capability of our approach in performing precise table manipulations.


\subsection{Ablation Studies}
We conducted a series of ablation studies to dissect the specific contributions of each key component within the SheetBrain framework.
\paragraph{Ablations on Understanding and Validation Modules.} We assess the contributions of the understanding and validation modules. The results show that removing the understanding module leads to a 3.3\% performance drop on both RealHitBench and SheetBench. Similarly, disabling the validation module results in a comparable performance decline, confirming its critical role in verifying and correcting the execution process. When both modules are removed simultaneously, performance further degrades to 73.3\% on RealHitBench and 73.8\% on SheetBench, which confirms that these components offer complementary advantages and jointly enhance the system's overall accuracy.

\paragraph{Ablations on Components of Execution Module.} We analyze the internal components of the execution module in \ref{tab:ablation_execution}. With the full suite of custom sheet toolkit, our method achieves an accuracy of 79.1\% on SheetBench. When the \texttt{inspector} tool is removed, accuracy drops to 77.3\%, and it further decreases to 73.1\% when the \texttt{search} tool or all tools are removed. More importantly, when we replaced our code sandbox with a traditional neural calling (JSON-based) approach, the accuracy plummeted to 65.1\%. This stark difference powerfully demonstrates the overwhelming advantage of our code sandbox architecture and its integrated symbolic tool suite for handling data-intensive spreadsheet tasks.

\paragraph{Ablations on Serialization Strategies for Sheet Content.} to investigate the impact of serialization strategies for sheet content on the performance of SheetBrain, we evaluate a variety of serialization formats, grouped into two main categories: Markdown-based and HTML-based representations. The Markdown-based approaches include a plain markdown format without positional metadata, as well as a variant that incorporates explicit cell position information. In contrast, the HTML-based methods include a plain HTML table representation, an HTML structure that mimics markdown style with added cell positions, a format that adds both cell positions and colspan attributes, and finally, a variant that utilizes row tags and colspan, while omitting explicit column positions. Table~\ref{tab:encoding-accuracy} presents the performance comparison across these serialization schemes. The results indicate that HTML-based serializations perform slightly better than their Markdown-based counterparts. More notably, the inclusion of cell position information yields a significant performance improvement across both categories, highlighting the importance of spatial context in structured data understanding. Interestingly, among the HTML variants, the format employing row tags with colspan, but without explicitly specifying cell positions, outperforms the version that encodes precise cell positions. This suggests that providing high-level row structure and span information may be more beneficial than overly detailed positional encoding, possibly due to reduced noise or increased generalizability.

\begin{table}[t]
    \centering
    \small
    \renewcommand{\arraystretch}{1.2}
    \setlength{\tabcolsep}{6pt}
    \begin{tabular}{ccccc}
        \toprule
        \textbf{Understanding} & \textbf{Validation} & \textbf{Realhitbench} & \textbf{SheetBench} \\
        \midrule
        \ding{51} & \ding{51} & 78.3 & 80.3 \\
        \ding{55} & \ding{51} & 75.0 & 77.0 \\
        \ding{51} & \ding{55} & 76.7 & 77.0 \\
        \ding{55} & \ding{55} & 73.3 & 73.8 \\
        \bottomrule
    \end{tabular}
    \caption{Ablation study with different components removed. \ding{51} indicates the module is used; \ding{55} indicates it is removed. Results are shown as hit rate percentages.}
    \vspace{-4pt}
    \label{tab:understanding_ablation}
\end{table}



\begin{table}[t]
\centering
\begin{tabular}{lc}
\toprule
\textbf{Execution Method} & \textbf{Accuracy (\%)} \\
\midrule
\multicolumn{2}{l}{\textbf{Code Sandbox Approach}} \\
\quad Full Method (All Tools) & 79.1 \\
\quad \quad w/o Inspector Tool & 77.3 \\
\quad \quad w/o Search Tool & 73.1 \\
\quad \quad w/o All Tools & 73.1 \\
\midrule
\multicolumn{2}{l}{\textbf{Traditional Tool-Use Approach}} \\
\quad Neural Calling (JSON-based) & 65.1 \\
\bottomrule
\end{tabular}
\caption{Ablation study of the execution module's components on the SheetBench dataset. We compare our full method using a code sandbox against variants with specific tools removed, and against a traditional neural calling (JSON-based) approach. The results highlight the significant performance gains from our integrated tool suite and the code sandbox architecture for data-intensive tasks.}
\label{tab:ablation_execution}
\end{table}



\begin{table}[t]
\centering
\resizebox{0.47\textwidth}{!}{
\begin{tabular}{llc}
\toprule
\textbf{Encoding Type} & \textbf{Variant} & \textbf{Accuracy} \\
\midrule
\multirow{2}{*}{Markdown} 
  & Pure Markdown & 63.3 \\
  & With Cell Position & 75.0 \\
\midrule
\multirow{4}{*}{HTML}
  & Pure HTML & 59.7 \\
  & MD-like HTML + Cell Pos. & 76.3 \\
  & HTML + Colspan + Cell Pos. & 76.0 \\
  & HTML + Colspan + Row Tag & 
  76.7 \\
\bottomrule
\end{tabular}
}
\caption{Accuracy of different table encoding variants.}
\label{tab:encoding-accuracy}
\end{table}
\subsection{Qualitative Analysis and Insights}
\paragraph{Strategy Adaptation Based on Table and Query Characteristics.} 
To gain a deeper understanding of SheetBrain's performance advantages and internal mechanisms, we conduct a qualitative analysis, which reveals a complex trade-off between symbolic computation and neural reasoning.

\begin{itemize}[leftmargin=*]
    \item \textbf{Symbolic computation} excels in scenarios involving large or extra-large tables and tasks requiring complex, multi-step calculations. For example, when a task requires filtering a 100,000-row table and then computing a conditional average, the neural approach is infeasible. The entire dataset cannot be encoded into the context. SheetBrain's symbolic approach handles this gracefully by loading the data into a pandas dataframe and executing a few lines of code, demonstrating a significantly higher performance ceiling for data-intensive operations.
    \item \textbf{Neural reasoning} also shows competitive potential in specific niche cases: small-to-medium-sized tables with complex hierarchical structures. When the entire table can be fully encoded within the LLM's context window, the model can sometimes leverage its powerful pattern-matching capabilities to directly comprehend a complex layout (e.g., intricate multi-level headers) holistically. For instance, Case 4 perfectly illustrates this niche scenario: the symbolic approach failed on a simple query about fishery landings because it overlooked the crucial \textbf{``of which'' breakdown items} within the complex row headers, leading to an incorrect parse. By contrast, neural reasoning, with the entire table encoded in its context, leveraged its holistic pattern-matching capabilities to directly comprehend this indented layout and accurately extract the answer. In such a scenario, a direct neural inference can be more straightforward than a step-by-step symbolic process of inspecting and parsing the structure through code.
\end{itemize}

The observation suggests that guiding the agent to differentiate its strategy based on the table characteristics and query type can lead to performance gains. Specifically, prompting the agent to output its thought process for small-to-medium tables with complex hierarchies, while defaulting to executing code for large tables or complex multi-step calculations, improves overall QA accuracy. This dynamic, scenario-based strategy selection allows the agent to better leverage the respective strengths of both neural reasoning and symbolic computation.

\paragraph{Limitations of Spreadsheet Previews in Existing Agents.} Current agents like ChatGPT~\cite{openai2023chatgpt} often struggle with spreadsheet analysis because they rely on a single, limited method for previewing data. By only loading and displaying the first few rows (e.g., using \texttt{df.head()}), they get a severely restricted view of the sheet's structure. This blind approach is a major limitation, especially when dealing with spreadsheets that have complex hierarchical layouts or multiple tables. As demonstrated in case analysis 1, this can lead to a fundamental misinterpretation of the data, such as failing to correctly identify a non-standard header row. This initial misunderstanding can cause the entire analysis to fail. This very observation, confirmed by poor performance on complex and multi-table categories in tests like SheetBench, highlights the critical need for a more comprehensive understanding module that can provide a full content overview before attempting any analysis.

\paragraph{The Need for Global Verification During Execution.} Our error analysis revealed that even symbolic methods have specific failure patterns. During the code execution and output observation process, the agent can become overly focused on the local information returned from a single step, causing it to ignore the global context. For instance, as shown in Case 3, when tasked with summing landings data, the agent correctly extracted numerical values from the sheet but failed to recognize the hierarchical structure, leading to a significant double-counting error by including both parent rows and their "of which" sub-categories in the total. This finding underscores the necessity of a global verification step and ultimately led us to incorporate a validation module. This module acts as a quality assurance and self-correction mechanism, prompting the agent to rigorously evaluate its entire process against the initial data structure before producing the final answer, thereby ensuring high reliability by catching and rectifying such logical flaws.

\begin{table}[t]
\centering
\resizebox{\linewidth}{!}{%
\begin{tabular}{lccccc}
\toprule
\textbf{Model} & \textbf{Complex} & \textbf{Multi-table} & \textbf{Large Sheet} & \textbf{Edit} & \textbf{Total} \\
\midrule
\textbf{Test Cases} & 21 & 20 & 20 & 8 & 69 \\
\midrule
\multicolumn{6}{l}{\textit{Vanilla Models}} \\
4o             & 16 & 16 & 2  & NA & 34 \\
o3                & 19 & 15 & 4  & NA & 38 \\
o4-mini           & 16 & 16 & 3  & NA & 35 \\
o3-mini           & 16 & 17 & 4  & NA & 37 \\
4.1               & 16 & 17 & 1  & NA & 34 \\
Qwen-3-32b        & 17 & 16 & 1  & NA & 34 \\
Deepseek-R1       & 16 & 17 & 2  & NA & 35 \\
\midrule
BizChat Analyst   & 18 & 14 & 9  & 6  & 47 \\
ChatGPT (4o)      & 13 & 8  & 9  & NA & 30 \\
StructGPT (4.1)   &  12   &  1  &  0  & NA &  13   \\
SheetAgent (4.1)  &   11  &  10  &  10  & 4   &  35   \\
SheetBrain (4.1)  & 20 & 18 & 11 &  6  & 55 \\
\bottomrule
\end{tabular}%
}
\caption{Sheetbench Results.}
\label{tab:model-comparison}
\end{table}

\subsection{SheetBench Analysis}

We evaluate a total of 12 models, grouped into vanilla LLMs and spreadsheet agents, on our proposed benchmark, SheetBench. The benchmark comprises 69 challenging tasks across four categories: Complex Tables, Multi-table Layouts, Large Sheets, and Editing Operations. Each model is evaluated on the same set of cases to ensure fairness and comparability. In addition, we provide a standardized prompting-based evaluator to assess QA outputs in a consistent and reproducible manner. The detailed performance breakdown is shown in Table~\ref{tab:model-comparison}. SheetBrain achieves the highest total score of 55 out of 69, significantly surpassing all other evaluated models across nearly every category. It demonstrates especially strong performance on the Complex and Multi-table subsets, achieving 20 out of 21 and 18 out of 20 respectively. These results highlight its superior capability in handling hierarchical structures, interleaved tables, and messy layouts that frequently appear in real-world spreadsheets.

\section{Related Work}

\subsection{LLM Prompting and Fine-tuning for Tabular Data}

LLMs have shown increasing promise for reasoning over structured tabular data, including question answering, semantic parsing, and data generation~\cite{fang2024large}. Early work demonstrates that general-purpose LLMs such as GPT-3~\cite{brown2020language} can answer table-based queries through in-context prompting~\cite{chen2022large}. To enhance compositional reasoning, DATER~\cite{ye2023large} decomposes both queries and tables into semantic units. TAPEX~\cite{liu2021tapex} advances table pretraining by fine-tuning a neural SQL executor on synthetic query–answer pairs, achieving state-of-the-art results on multiple benchmarks. Reasoning strategies have also evolved: chain-of-thought prompting~\cite{wei2022chain} enables models to perform symbolic reasoning step by step, while self-consistency decoding~\cite{wang2022self} improves accuracy by aggregating diverse reasoning paths. Binder~\cite{Binder} shows that Codex~\cite{chen2021evaluating} can generate SQL programs from natural language, demonstrating strong semantic parsing capabilities. StructGPT~\cite{jiang2023structgpt} builds on this by introducing a hierarchical reasoning framework tailored to structured tables. Despite these advances, most approaches are optimized for clean, single-table inputs and struggle with the complex, irregular, and layout-heavy structures common in real-world spreadsheets.

\subsection{Spreadsheet Agents}

Frameworks such as ReAct~\cite{yao2023react} have promoted recent efforts to move beyond static prompting and toward interactive spreadsheet agents~\cite{krishnan2025ai}. SheetCopilot~\cite{li2023sheetcopilot} formulates spreadsheet interaction as a software agent task, using a finite state machine and atomic operations to support task execution across a large benchmark suite. SheetAgent~\cite{chen2024sheetagent} incorporates planning, summarization, and code retrieval mechanisms to enable long-horizon reasoning over spreadsheets. SheetMind~\cite{zhu2025sheetmind} extends this line with a multi-agent architecture that decomposes user commands, translates intentions into formal grammar-based plans, and performs actions in real-time using the Google Sheets API. These agents represent a shift toward autonomous spreadsheet reasoning, yet many of them still operate on simplified inputs or assume clean table boundaries. Challenges such as imprecise user queries and mixed data types remain significant obstacles for robust deployment in enterprise or messy real-world spreadsheets.

\section{Conclusion}

In this work, we introduce SheetBrain, a neuro-symbolic agent based on a novel understand-execute-validate framework for reasoning over complex, large-scale spreadsheets. SheetBrain combines a deep understanding module, a symbolic execution engine operating within a lightweight code sandbox, and a self-correcting validation module to address intricate spreadsheet tasks that exceed the capabilities of purely neural models. In addition, we present SheetBench, a new benchmark comprising realistic spreadsheet scenarios with multi-table layouts, hierarchical structures, and large data volumes. Experimental results demonstrate that SheetBrain consistently outperforms strong vanilla LLMs from OpenAI, Qwen, and DeepSeek, as well as existing spreadsheet agents, across both public datasets and SheetBench. Our findings highlight the importance of symbolic dataflow for scalable and precise computation, and show that deep structural understanding is essential for mitigating the limited visibility and context fragmentation that hinder current LLM-based spreadsheet reasoning approaches.

\bibliography{aaai2026}
\appendix
\onecolumn

\section{\huge Case Analysis}

\subsection{Case 1: The Importance of a Content Overview}

This example demonstrates the critical difference between a blind and an informed approach to data analysis. When ChatGPT's data analysis feature was used without an overview, it simply loaded the first five rows. This led to a misinterpretation of the header, which caused the entire analysis to fail. In contrast, Sheetbrain, with its overview of the sheet's content, was able to correctly identify the absence of a standard header row. This foundational understanding allowed it to adapt its strategy and successfully solve the problem. This highlights that a preliminary view of the data's structure is essential for accurate and effective analysis. \\ Query: How many users handled more than 11 chats serviced, and what is the total number of chats they serviced? \\ Answer: "2, 24"

\begin{figure}[htbp]
    \centering
    \includegraphics[width=\columnwidth]{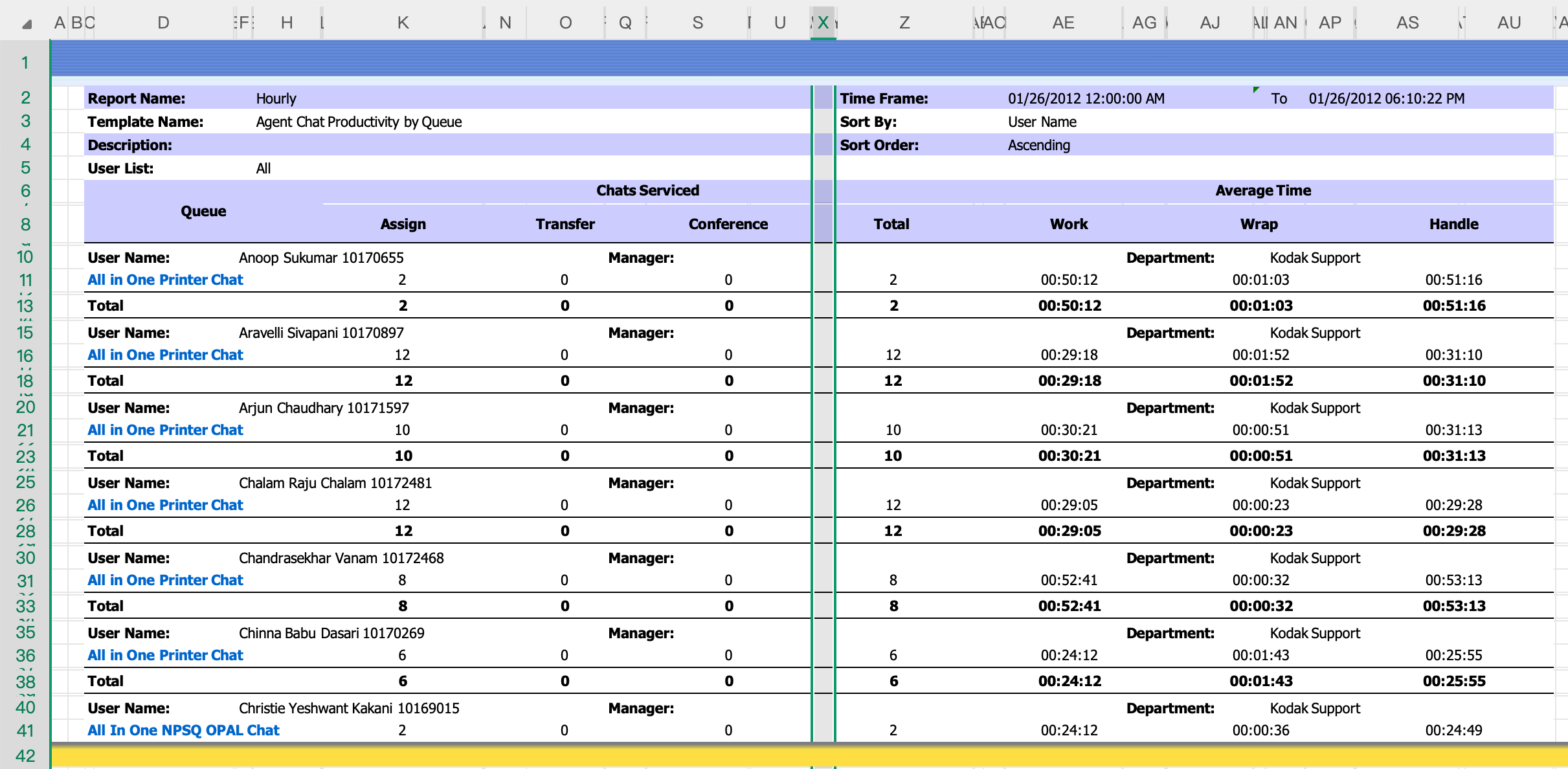}
    \caption{Case 1 Sheet}
    \label{fig:comparison}
\end{figure}

This example highlights the contrasting outcomes between a blind and an informed approach to data analysis.
When using a blind approach, as demonstrated by the ChatGPT data analysis feature, an initial misinterpretation of the data's structure (specifically, the absence of a standard header row) led to a failed analysis.
In contrast, an informed approach, like that used by Sheetbrain, first provided an overview of the sheet's content. This allowed it to correctly identify the structural anomaly and adapt its strategy, leading to a successful analysis. This underscores the crucial role of a preliminary data overview for effective problem-solving.

\begin{figure}[htbp]
    \centering
    \begin{minipage}{0.48\columnwidth}
        \includegraphics[width=\columnwidth]{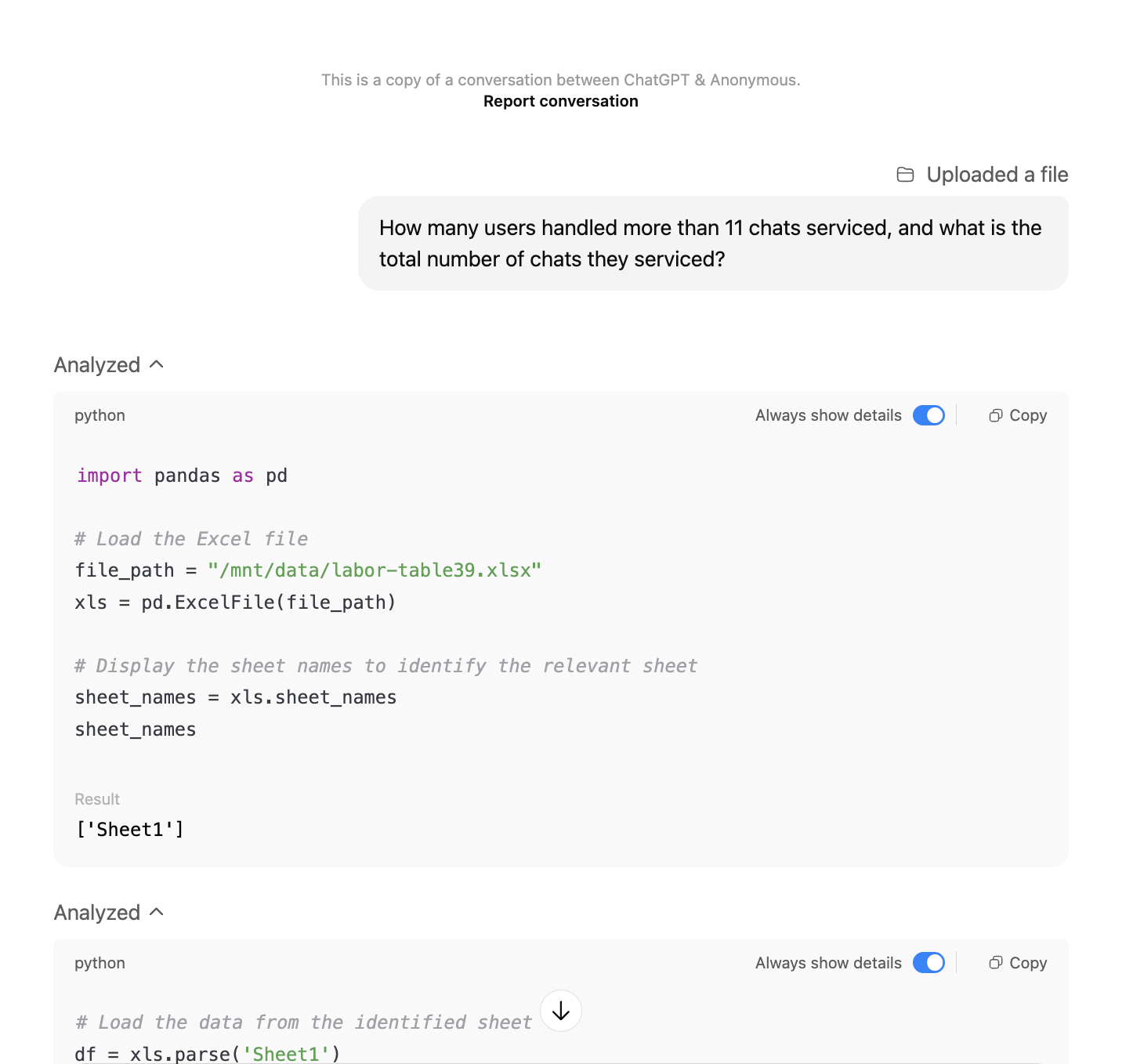}
        \caption*{(a) First step}
    \end{minipage}
    \hfill
    \begin{minipage}{0.48\columnwidth}
        \includegraphics[width=\columnwidth]{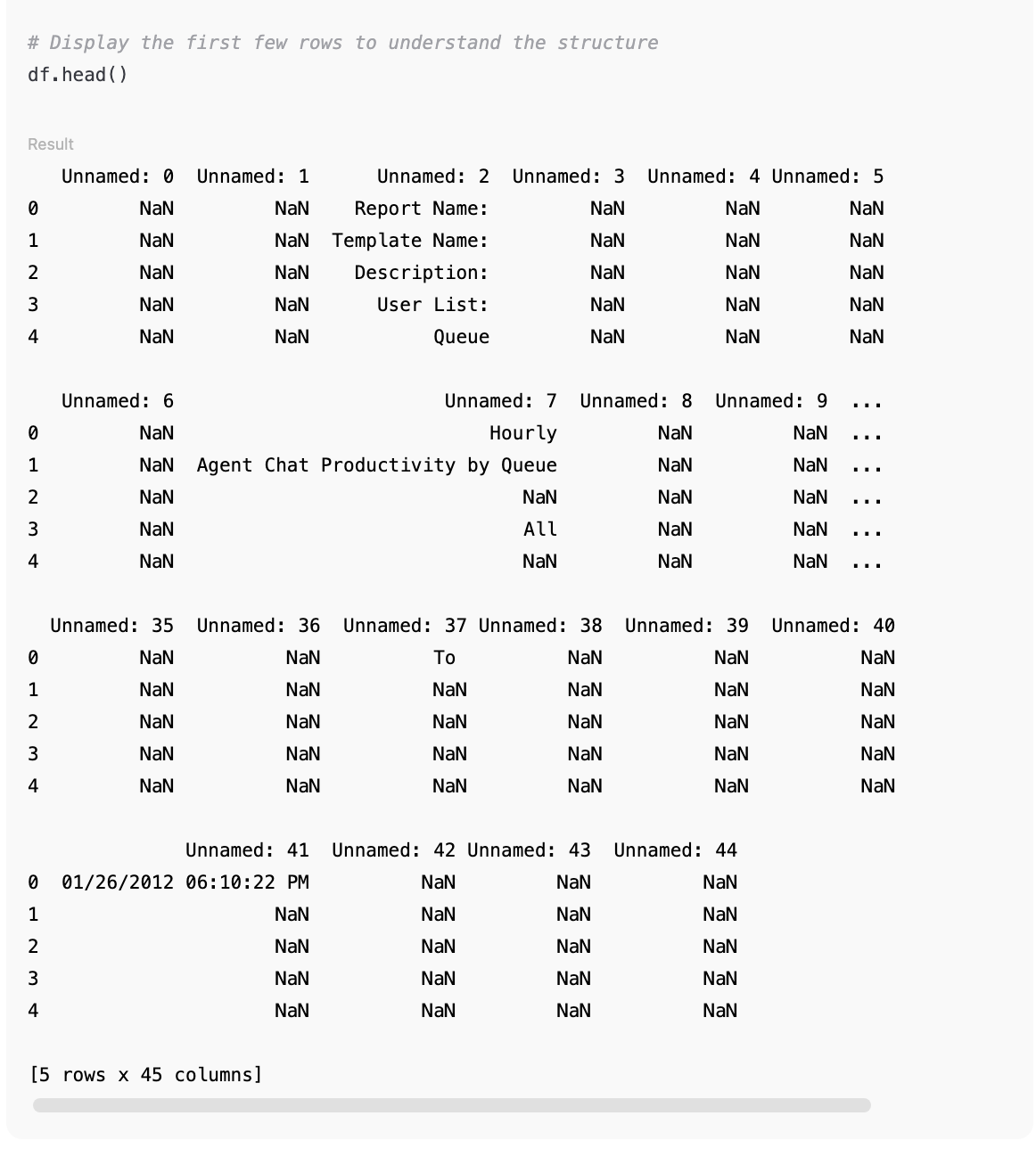}
        \caption*{(b) Second step}
    \end{minipage}
    \begin{minipage}{0.48\columnwidth}
        \includegraphics[width=\columnwidth]{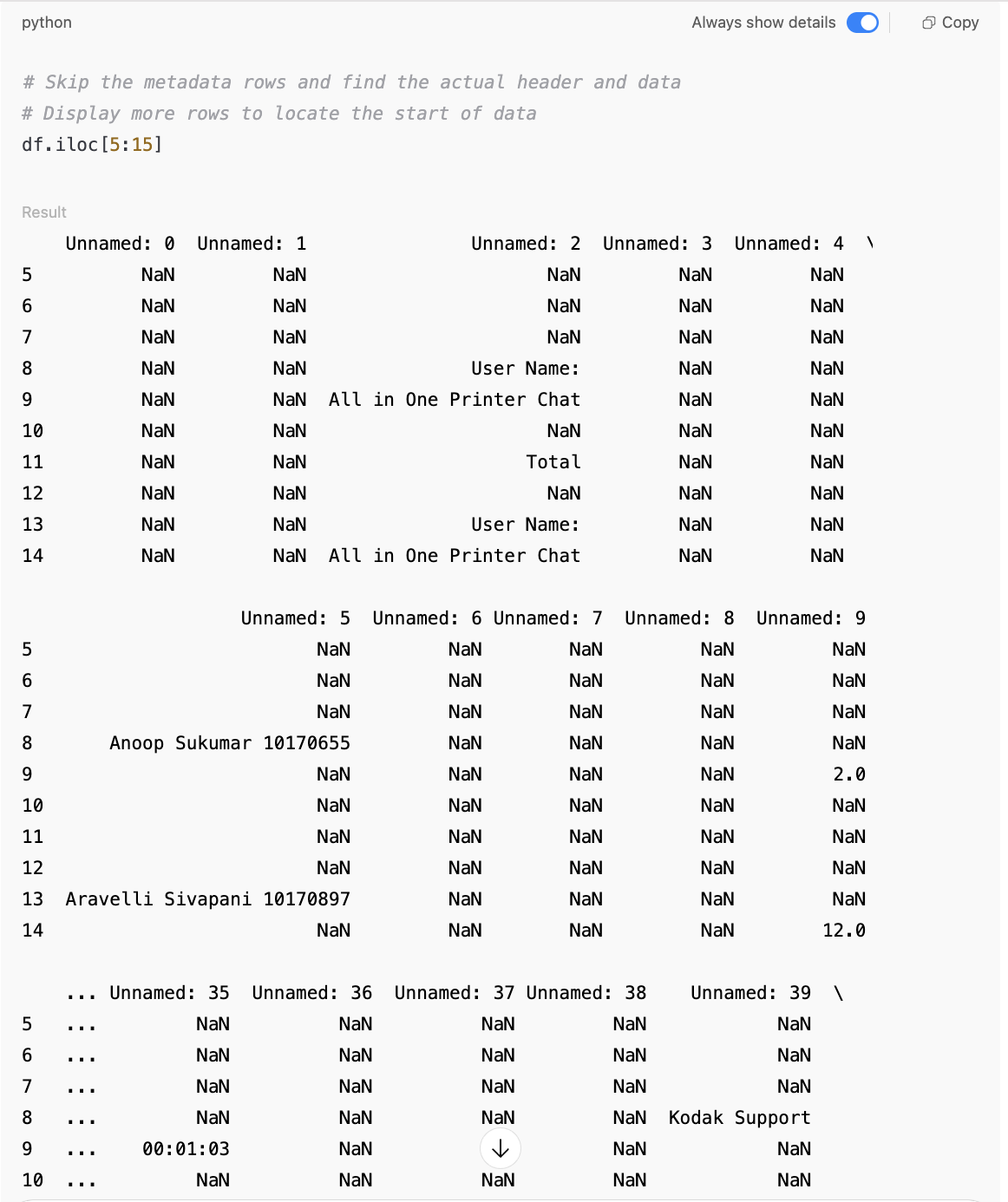}
        \caption*{(c) Third step}
    \end{minipage}
    \hfill
    \begin{minipage}{0.48\columnwidth}
        \includegraphics[width=\columnwidth]{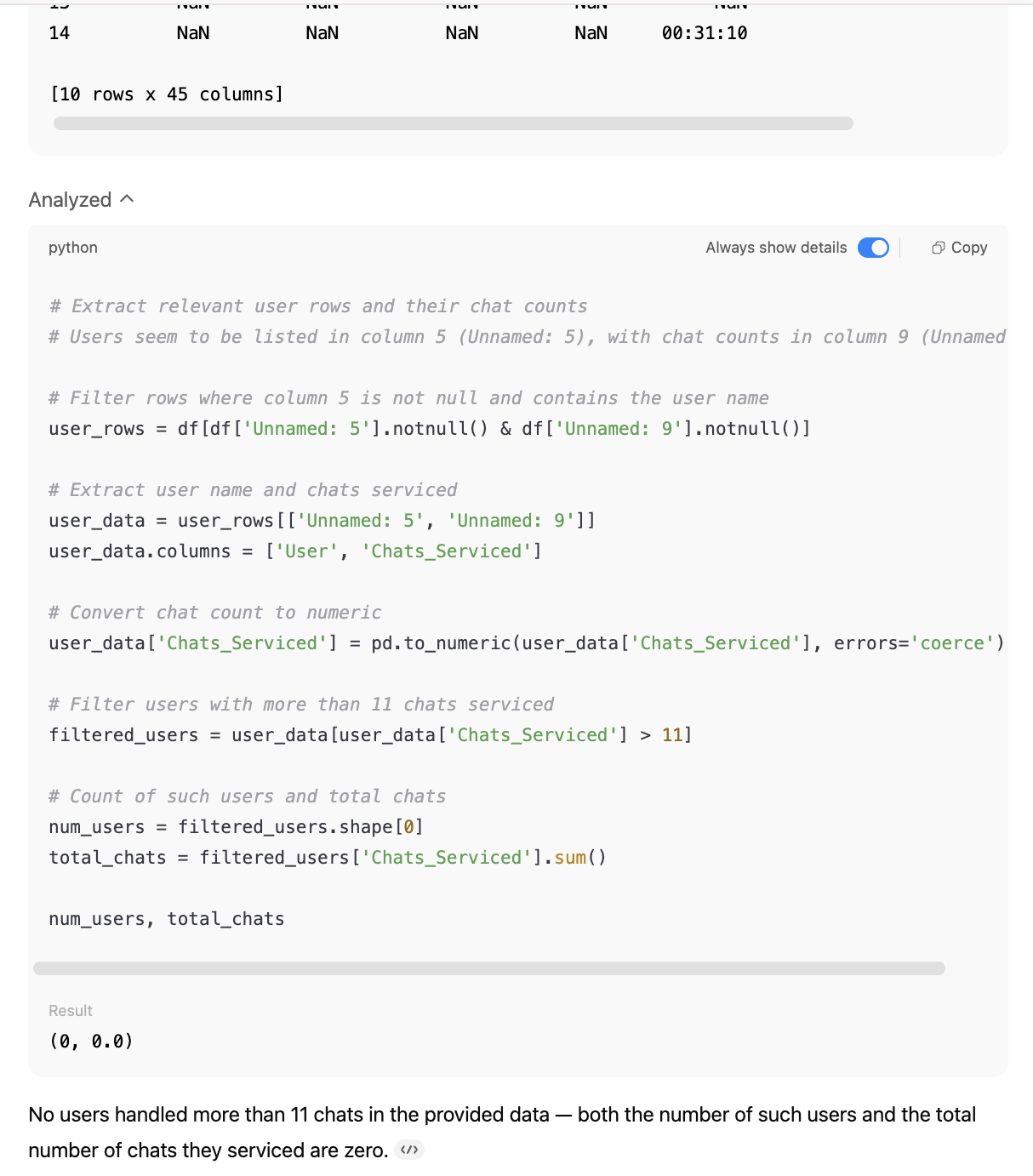}
        \caption*{(d) Final step}
    \end{minipage}
    \caption{ChatGPT's blind approach, demonstrating the failed analysis at each stage.}
    \label{fig:comparison}
\end{figure}

\subsubsection{SheetBrain: Correct Approach}
Given the absence of a standard header row in the sheet, it is necessary to utilize `header\_row=0` to ensure that all rows are treated as data. The data will be reloaded with `header\_row=0`, and the analysis will subsequently be repeated.

\begin{Verbatim}
# Load the sheet as a DataFrame, treating all rows as data (no header)
df = get_sheet_as_dataframe('Sheet1', header_row=0)

# Identify all rows where column C contains "User Name:" to delineate user blocks
user_rows = df[df == "User Name:"].index.tolist()

# For each identified user, locate the subsequent "Total" row following their "User Name:" row
results = []
for idx, user_row in enumerate(user_rows):
    # Extract the user name from column F
    user_name = df.iloc[user_row, 5]
    # Search for the next "Total" row in column C subsequent to user_row
    if idx < len(user_rows) - 1:
        # Limit the search to the extent of the next user block
        block = df.iloc[user_row+1:user_rows[idx+1]]
    else:
        # For the final user, extend the search to the end of the data
        block = df.iloc[user_row+1:]
    total_row = block[block == "Total"]
    if not total_row.empty:
        # Retrieve the first "Total" row within the block
        total_idx = total_row.index
        # Obtain the total number of chats serviced from column K (index 10)
        try:
            chats = int(df.iloc[total_idx, 10])
        except (ValueError, TypeError):
            chats = 0
        results.append({'user': user_name, 'chats': chats})

# Filter users who have serviced more than 11 chats
over_11 = [r for r in results if r['chats'] > 11]
num_users = len(over_11)
total_chats = sum(r['chats'] for r in over_11)

# Display the identified users and their respective chat counts for clarity
over_11, num_users, total_chats

Code execution result:
Expression result: ([{'user': 'Aravelli Sivapani 10170897', 'chats': 12}, 

{'user': 'Chalam Raju Chalam 10172481', 'chats': 12}], 2, 24)
\end{Verbatim}

\subsection{Case 2: Agent Symbolic Dataflow vs. Neural Dataflow}

This example demonstrates the key differences between a Symbolic Dataflow and a Neural Dataflow approach in handling a data analysis query on a large dataset.

Query: Which issuing country has the highest number of transactions? The answer must be just the country code. If a question does not have a relevant or applicable answer for the task, please respond with 'Not Applicable'.

Expected Answer: "NL"

\begin{figure}[htbp]
    \centering
    \includegraphics[width=\columnwidth]{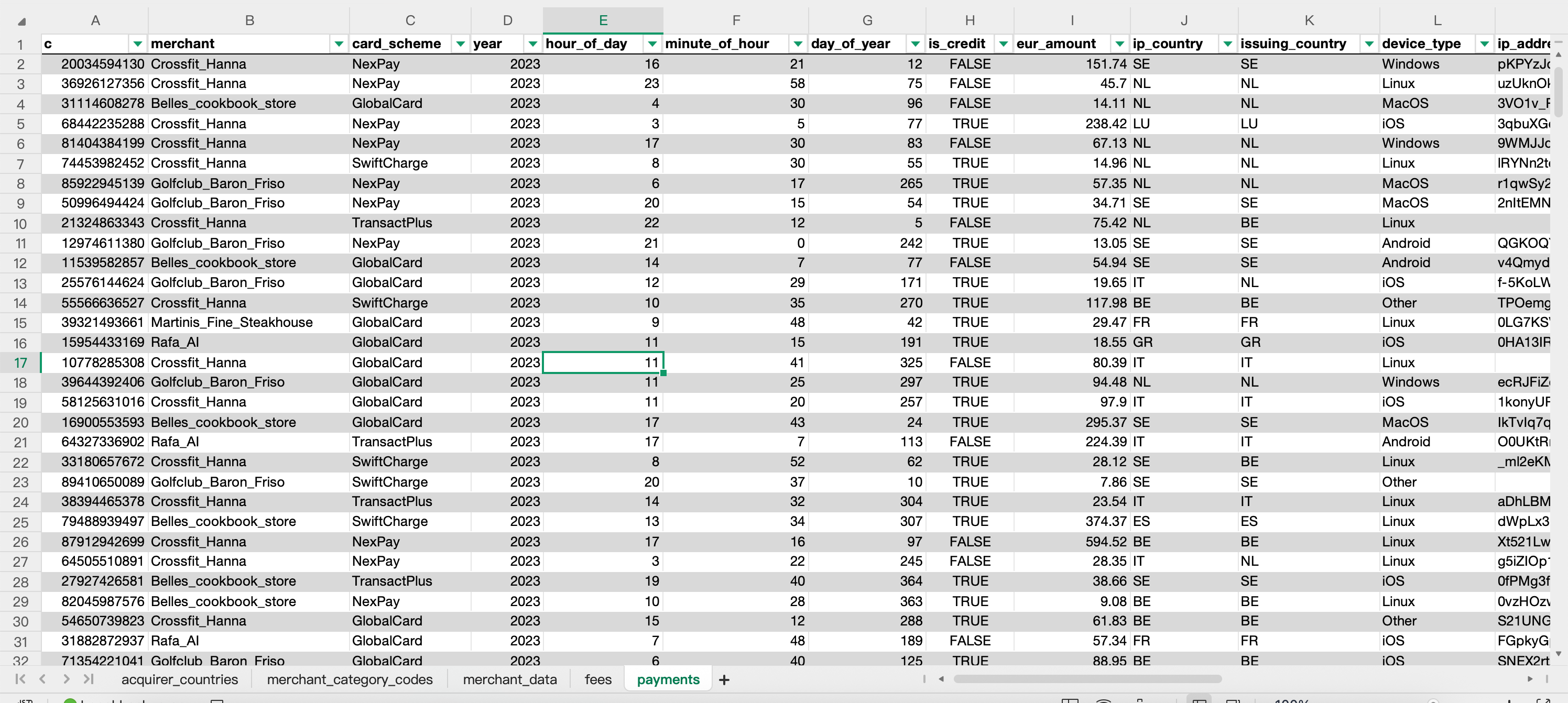}
    \caption{Case 2 Sheet}
    \label{fig:comparison}
\end{figure}

\subsubsection{Symbolic Dataflow: Correct Approach}
Symbolic Dataflow uses code variables to handle data precisely. The output of each step is stored in a named variable (e.g., df), and the agent writes code that directly references these variables. This method is highly effective for complex computations and large datasets, as it avoids the context limitations of text-based systems. As shown below, it correctly identifies the full dataset's structure and counts all transactions to find the country with the most transactions.

\begin{Verbatim}
# Step 1: Load only the 'issuing_country' column from the 'payments' sheet for efficiency
import pandas as pd

# Since the sheet is very large, let's load only the header row to get column names
df_head = get_sheet_as_dataframe("payments", header_row=1, max_rows=1)
issuing_country_col = [col for col in df_head.columns if "issuing_country" in col][0]

# Now, load only the 'issuing_country' column for all rows
# We'll use chunks to avoid memory issues if needed, but let's try loading the column 
directly first
df = get_sheet_as_dataframe("payments", header_row=1, max_rows=None)[[issuing_country_col]]

# Count occurrences of each country code
country_counts = df[issuing_country_col].value_counts()

# Print the top result
print(country_counts.head(3))
\end{Verbatim}

\textbf{Execution result:}
\begin{Verbatim}
Output:
issuing_country
NL     29622
IT     28329
BE     23040
Name: count, dtype: int64

Final Answer: NL
\end{Verbatim}

\subsubsection{Neural Dataflow: Failed Approach}
Neural Dataflow is limited by its reliance on text descriptions. It has to feed tool outputs back into the context, which is constrained by the model's token limit. This makes it unsuitable for extensive operations, where the context can quickly become overwhelmed. In this example, the agent attempts to inspect only a small sample (K2:K1001) of the data due to these limitations. Based on this incomplete and biased sample, it incorrectly concludes that "SE" is the answer, failing the task.

\begin{Verbatim}
[Agent] Thinking...
[Parse] Thought: 'To determine which issuing country has the highest number of transactions, 
I need to count the occurrences of each country code in the "issuing_country" column in the 
'payments' sheet. Since the sample is inadequate, I need to perform a keyword search
for all unique issuing countries. 
I will first inspect the relevant column to get a list of countries and then count their occurrences.'
[Parse] Action: {'tool_name': 'inspector', 'parameters': {'range_references': ['K2:K1001'], 'sheet_name': 'payments'}}
 [LLM Action] Preparing to call tool: inspector, parameters: {'range_references': ['K2:K1001'], 'sheet_name': 'payments'}
[Tool: inspector] Called with ranges ['K2:K1001'], sheet 'payments', attributes 'None'
[Tokenizer] Using character approximation (3 chars = 1 token)
[Table Token Count] Table Token length: 1749 tokens (approx from 6999 chars)
[Tool Tracking] Tools used so far: ['inspector']
[Agent Observation] Tool inspector returned result:
**Multiple Ranges Content in sheet 'payments':**

**Range 1 (K2:K1001):**
| SE |
| NL |
| NL |
| LU |
| NL |
| NL |
| NL |
| SE |
| BE |
| SE |
| SE |
| NL |
... (truncated output for brevity) ...

---------- TURN 3 ----------
[Agent] Thinking...
[Parse] Found Final Answer: 'SE'

================== FINAL ANSWER ==================
SE
\end{Verbatim}

\subsection{Case 3: Necessary Verification
}

In this case study, we observe a common pitfall: the agent, while executing code, becomes overly focused on the local information returned by the code, neglecting the broader, global context provided at the outset. This narrow perspective resulted in a significant error, for instance, double-counting during summation tasks.
To address this, we implemented a verification module just before the final output. This module prompts the agent to reflect on its entire process, re-examining the initial global information as well as the steps taken during execution. Through this self-Validation, SheetBrain successfully identified and corrected its previous mistake, ultimately providing the accurate result and resolving the double-counting issue. This success highlights the critical role of a verification step in preventing errors and ensuring accuracy, particularly when agents are prone to losing sight of the bigger picture.

 "question": "What is the total landings (tonnes live weight) for Scotland in 2023, and how does it compare to the total landings for England, Wales, and N.I.? "
 "expected answer": "163802, 6987, much higher",

\begin{figure}[htbp]
    \centering
    \includegraphics[width=\columnwidth]{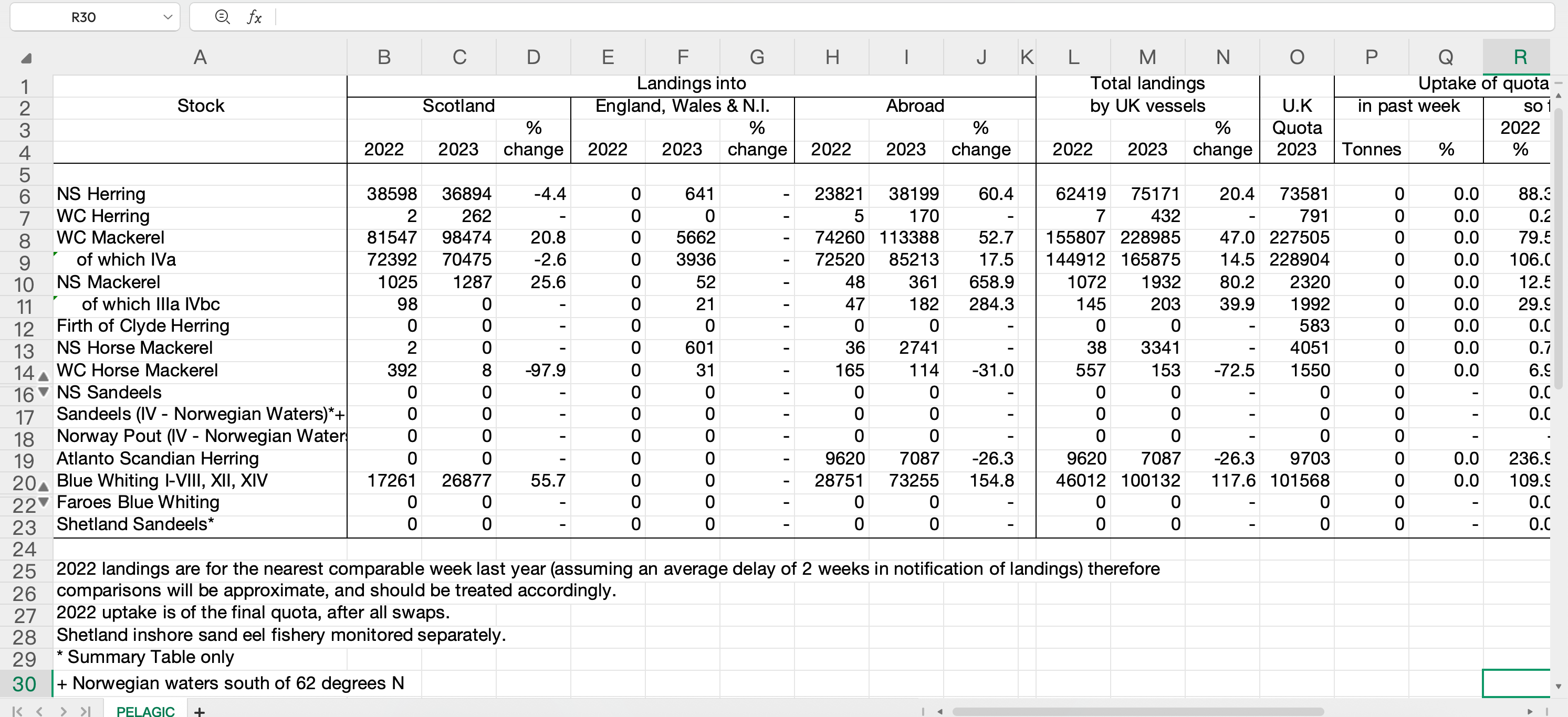}
    \caption{Case 3 Sheet}
    \label{fig:comparison}
\end{figure}

\subsubsection{SheetBrain without validator: Wrong}

\begin{Verbatim}
 [ITERATION 1] EXECUTION MODULE
----------------------------------------
 [ExecutionModule] Starting multi-turn analysis for: 'What is the total landings (tonnes live weight) for Scotland in 2023, and how does it compare to the total landings for England, Wales, and N.I.?'

---------- EXECUTION TURN 1 ----------
 [ExecutionModule] Agent thinking...
 [ExecutionModule] Executing Python code:
# Step 1: Explore the relevant columns and rows for the question.
# The question asks for "total landings (tonnes live weight)" for Scotland and for England, Wales, and N.I. in 2023.

# From the preview:
# - Scotland 2023: Column C ("C4:2023" under "B2:Scotland")
# - England, Wales & N.I. 2023: Column F ("F4:2023" under "E2:England, Wales & N.I.")

# The "Total landings by UK vessels" row is likely the summary row for all stocks.
# Let's search for the row label that indicates the total for Scotland and for England, Wales & N.I.

# Let's inspect the first 20 rows in columns A to N to find the relevant "Total" row.
data = inspector("A1:N20", "PELAGIC")
for i, row in enumerate(data):
    print(f"Row {i+1}: {row}")
 [ExecutionModule] Execution result:
Output:
Row 1: [None, 'Landings into', None, None, None, None, None, None, None, None, None, 'Total landings', None, None]
Row 2: ['Stock', 'Scotland', None, None, 'England, Wales & N.I.', None, None, 'Abroad', None, None, None, 'by UK vessels', None, None]
Row 3: [None, None, None, '%', None, None, '%', None, None, '%', None, None, None, '%']
Row 4: [None, 2022, 2023, 'change', 2022, 2023, 'change', 2022, 2023, 'change', None, 2022, 2023, 'change']
Row 5: [None, None, None, None, None, None, None, None, None, None, None, None, None, None]
Row 6: ['NS Herring', 38597.9, 36893.62, -4.41547338067615, 0, 640.896, '-', 23821.22, 38198.5493, 60.3551342038737, None, 62419.12, 75171.0553, 20.4295339312698]
Row 7: ['WC Herring', 1.78, 262.09, '-', 0, 0, '-', 5.35, 170.233, '-', None, 7.13, 432.323, '-']
Row 8: ['WC Mackerel', 81546.81, 98473.65, 20.757206811646, 0, 5662.28299999999, '-', 74260.35, 113387.945, 52.6897530108597, None, 155807.16, 228985.001, 46.9669307880331]
Row 9: ['    of which IVa', 72391.79, 70475.45, -2.64717863724597, 0, 3935.516, '-', 72519.96, 85213.217, 17.5031218991296, None, 144911.75, 165874.81, 14.4660871185394]
Row 10: ['NS Mackerel', 1024.67, 1287.44, 25.6443537919526, 0, 52.293, '-', 47.6, 361.234, 658.894957983193, None, 1072.27, 1932.077, 80.1856808453095]
Row 11: ['     of which IIIa IVbc', 97.74, 0, '-', 0, 21.304, '-', 47.25, 181.583, 284.302645502646, None, 144.99, 202.887, 39.9317194289262]
Row 12: ['Firth of Clyde Herring', 0, 0, '-', 0, 0, '-', 0, 0, '-', None, 0, 0, '-']
Row 13: ['NS Horse Mackerel', 1.5, 0, '-', 0, 600.66, '-', 36.01, 2740.804, '-', None, 37.51, 3341.464, '-']
Row 14: ['WC Horse Mackerel', 392.27, 8.36, -97.8688148469167, 0, 30.816, '-', 165.05, 113.967, -30.9500151469252, None, 557.32, 153.143, -72.5215316155889]
Row 15: ['NS Blue Whiting EU', 0, 0, '-', 0, 0, '-', 3.21, 0, '-', None, 3.21, 0, '-']
Row 16: ['NS Sandeels', 0, 0, '-', 0, 0, '-', 0, 0, '-', None, 0, 0, '-']
Row 17: ['Sandeels (IV - Norwegian Waters)*+', 0, 0, '-', 0, 0, '-', 0, 0, '-', None, 0, 0, '-']
Row 18: ['Norway Pout (IV - Norwegian Waters)*+', 0, 0, '-', 0, 0, '-', 0, 0.141, '-', None, 0, 0.141, '-']
Row 19: ['Atlanto Scandian Herring', 0, 0, '-', 0, 0, '-', 9620.33, 7087.29, -26.3300739163833, None, 9620.33, 7087.29, -26.3300739163833]
Row 20: ['Blue Whiting I-VIII, XII, XIV', 17261.3, 26877.05, 55.7069861482044, 0, 0, '-', 28750.62, 73254.908, 154.794185308004, None, 46011.92, 100131.958, 117.621777139489]



---------- EXECUTION TURN 2 ----------
 [ExecutionModule] Agent thinking...
==================================================
I found that the data is organized by stock, with each row representing a different stock (e.g., "NS Herring", "WC Herring", etc.), and the columns represent landings for Scotland and for England, Wales & N.I. in 2022 and 2023. 

- Scotland 2023 values are in column C (index 2 in Python, "C" in Excel).
- England, Wales & N.I. 2023 values are in column F (index 5 in Python, "F" in Excel).

There is no explicit "Total" row for Scotland or England, Wales & N.I. in the first 20 rows. Instead, the totals must be calculated by summing the 2023 values for each stock in the relevant columns.

Let me sum the 2023 landings for Scotland (column C) and for England, Wales & N.I. (column F) for all stocks listed in the table (likely rows 6 to 23, based on the preview), and compare the results.
==================================================

---------- EXECUTION TURN 3 ----------
 [ExecutionModule] Agent thinking...
==================================================
 [ExecutionModule] Executing Python code:
# Sum the 2023 landings for Scotland (column C) and England, Wales & N.I. (column F)
# Stocks appear to be listed from row 6 to row 23 (A6:A23)
import numpy as np

# Get the relevant data range: rows 6-23, columns A-F (to include stock name and both 2023 columns)
data = inspector("A6:F23", "PELAGIC")

scotland_2023_total = 0.0
ewni_2023_total = 0.0

for row in data:
    stock = row[0]
    scotland_2023 = row[2]
    ewni_2023 = row[5]
    # Only sum if the value is a number (not '-', None, or text)
    try:
        if scotland_2023 not in [None, '-', '']:
            scotland_2023_total += float(scotland_2023)
    except:
        pass
    try:
        if ewni_2023 not in [None, '-', '']:
            ewni_2023_total += float(ewni_2023)
    except:
        pass

print(f"Scotland 2023 total landings: {scotland_2023_total:.2f} tonnes")
print(f"England, Wales & N.I. 2023 total landings: {ewni_2023_total:.2f} tonnes")
print(f"Difference (Scotland - E,W&NI): {scotland_2023_total - ewni_2023_total:.2f} tonnes")
 [ExecutionModule] Execution result:
Output:
Scotland 2023 total landings: 261154.71 tonnes
England, Wales & N.I. 2023 total landings: 10943.77 tonnes
Difference (Scotland - E,W&NI): 250210.94 tonnes



---------- EXECUTION TURN 4 ----------
 [ExecutionModule] Agent thinking...
==================================================
Final Answer:  
The total landings (tonnes live weight) for Scotland in 2023 are **261,154.71 tonnes**, while for England, Wales, and N.I. the total is **10,943.77 tonnes**. Scotland's total landings exceed those of England, Wales, and N.I. by **250,210.94 tonnes**.


==================================================
==================================================

\end{Verbatim}

\subsubsection{SheetBrain with validator:
Correct}

\begin{Verbatim}
**VALIDATION_STATUS:** FAILED

**CONFIDENCE_SCORE:** 0.4

**ISSUES_FOUND:**
- The agent summed all rows in the Scotland_2023 and EWNI_2023 columns without accounting for hierarchical relationships in the data (e.g., "of which" rows, which are subcategories of the main stock above them).
- This approach results in double-counting for stocks that have both a parent and "of which" sub-rows (e.g., "WC Mackerel" and "of which IVa").
- The agent did not filter out subtotal or breakdown rows (such as "of which IVa", "of which IIIa IVbc", etc.), which should not be summed together with their parent rows.
- There was no explicit check or use of the inspector/inspector_attribute to verify the editing area before giving the final answer, as required by the instructions.
- The agent did not provide a breakdown or verification of which rows were included in the sum, making it difficult to audit the calculation.
- The answer format is otherwise clear and directly addresses the user's question, but the calculation method is flawed.

**IMPROVEMENT_FEEDBACK:**
- Before summing, use the inspector to review the "Stock" column and identify which rows are parent categories and which are subcategories (e.g., rows starting with "of which", "of which IVa", "of which IIIa IVbc", etc.).
- Exclude all "of which" and similar breakdown rows from the sum to avoid double-counting. Only sum the main stock rows.
- Consider using a filter such as: only include rows where the "Stock" value does NOT start with "of which" or is not an indented/breakdown row.
- Re-execute the sum for Scotland_2023 and EWNI_2023 using only the main stock rows.
- Use inspector or inspector_attribute to explicitly check the editing area and confirm which rows are being included in the calculation before providing the final answer.
- Provide a brief list of which stock names were included in the sum for transparency.

**FINAL_ASSESSMENT:**
The solution provides a clear and direct answer to the user's question, but the calculation method is incorrect due to double-counting caused by including both parent and subcategory ("of which") rows in the sum. This is a common pitfall in hierarchical data and leads to inflated totals. The agent should have filtered out breakdown rows and only summed main stock categories. The answer should be recalculated with proper data handling to ensure accuracy. Confidence is low until this is corrected.
==================================================
[ValidationModule] Validation completed.
[ValidationModule] Confidence Score: 0.40
[ValidationModule] Validation: FAILED
[ValidationModule] Issues found - recommending re-execution
[ITERATION 1] Validation completed in 6.97s
[ITERATION 1] Confidence: 0.40
[ITERATION 1] Validation: FAILED
[CONTINUE] Issues found, preparing for iteration 2

[ITERATION 2/3] EXECUTE-REFLECT CYCLE
============================================================
[ITERATION 2] EXECUTION MODULE
----------------------------------------
[ExecutionModule] Starting multi-turn analysis for: 'What is the total landings (tonnes live weight) for Scotland in 2023, and how does it compare to the total landings for England, Wales, and N.I.?'

---------- EXECUTION TURN 1 ----------
[ExecutionModule] Agent thinking...
==================================================
[ExecutionModule] Executing Python code:
# Step 1: Use inspector to review the "Stock" column (A6:A23) and the corresponding 2023 columns for Scotland (C) and England, Wales & N.I. (F)
# We'll inspect a reasonable range to capture all relevant stock rows and their breakdowns.

stock_range = "A6:A23"
scotland_2023_range = "C6:C23"
ewni_2023_range = "F6:F23"

stock_names = inspector(stock_range, "PELAGIC")
scotland_2023 = inspector(scotland_2023_range, "PELAGIC")
ewni_2023 = inspector(ewni_2023_range, "PELAGIC")

# Print the stock names to identify which are main stocks and which are breakdowns ("of which", indented, etc.)
for i, row in enumerate(stock_names):
    print(f"Row {i+6}: {row[0]}")
[ExecutionModule] Execution result:
Output:
Row 6: NS Herring
Row 7: WC Herring
Row 8: WC Mackerel
Row 9:     of which IVa
Row 10: NS Mackerel
Row 11:      of which IIIa IVbc
Row 12: Firth of Clyde Herring
Row 13: NS Horse Mackerel
Row 14: WC Horse Mackerel
Row 15: NS Blue Whiting EU
Row 16: NS Sandeels
Row 17: Sandeels (IV - Norwegian Waters)*+
Row 18: Norway Pout (IV - Norwegian Waters)*+
Row 19: Atlanto Scandian Herring
Row 20: Blue Whiting I-VIII, XII, XIV
Row 21:    of which Bay of Biscay BW
Row 22: Faroes Blue Whiting
Row 23: Shetland Sandeels*



---------- EXECUTION TURN 2 ----------
[ExecutionModule] Agent thinking...
==================================================
LLM RESPONSE CONTENT:
==================================================
I found that the "Stock" column contains both main stock rows and breakdown rows (such as those starting with "of which" or indented). To avoid double-counting, we need to exclude the breakdown rows from our sum.

**Identified breakdown rows (to exclude):**
- Row 9:     of which IVa
- Row 11:      of which IIIa IVbc
- Row 21:    of which Bay of Biscay BW

**Main stock rows (to include):**
- Row 6: NS Herring
- Row 7: WC Herring
- Row 8: WC Mackerel
- Row 10: NS Mackerel
- Row 12: Firth of Clyde Herring
- Row 13: NS Horse Mackerel
- Row 14: WC Horse Mackerel
- Row 15: NS Blue Whiting EU
- Row 16: NS Sandeels
- Row 17: Sandeels (IV - Norwegian Waters)*+
- Row 18: Norway Pout (IV - Norwegian Waters)*+
- Row 19: Atlanto Scandian Herring
- Row 20: Blue Whiting I-VIII, XII, XIV
- Row 22: Faroes Blue Whiting
- Row 23: Shetland Sandeels*

Let me extract the corresponding Scotland 2023 and England, Wales & N.I. 2023 values for these main stock rows, and sum them for the final answer. I will also print the values for transparency.

```python
# Indices of main stock rows (0-based, relative to A6:A23)
main_stock_indices = [0, 1, 2, 4, 6, 7, 8, 9, 10, 11, 12, 13, 14, 16, 17]

included_stocks = []
scotland_2023_values = []
ewni_2023_values = []

for idx in main_stock_indices:
    stock = stock_names[idx][0]
    scot_val = scotland_2023[idx][0]
    ewni_val = ewni_2023[idx][0]
    included_stocks.append(stock)
    scotland_2023_values.append(scot_val)
    ewni_2023_values.append(ewni_val)
    print(f"{stock}: Scotland 2023 = {scot_val}, England, Wales & N.I. 2023 = {ewni_val}")

# Now, let's clean and sum the values (convert to float, treat '-' or empty as 0)
def safe_float(val):
    try:
        return float(val)
    except:
        return 0.0

scotland_2023_sum = sum(safe_float(x) for x in scotland_2023_values)
ewni_2023_sum = sum(safe_float(x) for x in ewni_2023_values)

print(f"\nTotal Scotland 2023: {scotland_2023_sum}")
print(f"Total England, Wales & N.I. 2023: {ewni_2023_sum}")
print(f"Difference (Scotland - EWNI): {scotland_2023_sum - ewni_2023_sum}")
```

==================================================
[ExecutionModule] Execution result:
Output:
NS Herring: Scotland 2023 = 36893.62, England, Wales & N.I. 2023 = 640.896
WC Herring: Scotland 2023 = 262.09, England, Wales & N.I. 2023 = 0
WC Mackerel: Scotland 2023 = 98473.65, England, Wales & N.I. 2023 = 5662.28299999999
NS Mackerel: Scotland 2023 = 1287.44, England, Wales & N.I. 2023 = 52.293
Firth of Clyde Herring: Scotland 2023 = 0, England, Wales & N.I. 2023 = 0
NS Horse Mackerel: Scotland 2023 = 0, England, Wales & N.I. 2023 = 600.66
WC Horse Mackerel: Scotland 2023 = 8.36, England, Wales & N.I. 2023 = 30.816
NS Blue Whiting EU: Scotland 2023 = 0, England, Wales & N.I. 2023 = 0
NS Sandeels: Scotland 2023 = 0, England, Wales & N.I. 2023 = 0
Sandeels (IV - Norwegian Waters)*+: Scotland 2023 = 0, England, Wales & N.I. 2023 = 0
Norway Pout (IV - Norwegian Waters)*+: Scotland 2023 = 0, England, Wales & N.I. 2023 = 0
Atlanto Scandian Herring: Scotland 2023 = 0, England, Wales & N.I. 2023 = 0
Blue Whiting I-VIII, XII, XIV: Scotland 2023 = 26877.05, England, Wales & N.I. 2023 = 0
Faroes Blue Whiting: Scotland 2023 = 0, England, Wales & N.I. 2023 = 0
Shetland Sandeels*: Scotland 2023 = 0, England, Wales & N.I. 2023 = 0

Total Scotland 2023: 163802.20999999996
Total England, Wales & N.I. 2023: 6986.947999999989
Difference (Scotland - EWNI): 156815.262



---------- EXECUTION TURN 3 ----------
[ExecutionModule] Agent thinking...
==================================================
LLM RESPONSE CONTENT:
==================================================
Final Answer:  
The total landings (tonnes live weight) for Scotland in 2023 are **163,802.21 tonnes**, while the total for England, Wales, and N.I. is **6,986.95 tonnes**. Scotland's total exceeds that of England, Wales, and N.I. by **156,815.26 tonnes**.

**Stocks included in the sum (main stocks only, breakdowns excluded):**
- NS Herring
- WC Herring
- WC Mackerel
- NS Mackerel
- Firth of Clyde Herring
- NS Horse Mackerel
- WC Horse Mackerel
- NS Blue Whiting EU
- NS Sandeels
- Sandeels (IV - Norwegian Waters)*+
- Norway Pout (IV - Norwegian Waters)*+
- Atlanto Scandian Herring
- Blue Whiting I-VIII, XII, XIV
- Faroes Blue Whiting
- Shetland Sandeels*

**No breakdown or "of which" rows were included, ensuring no double-counting.**


\end{Verbatim}

\subsection{Case 4: Neural Reasoning vs Symbolic Computation}

In scenarios involving complex spreadsheet structures, such as those with multi-level headers or multiple linked sheets, relying solely on Symbolic Computation can be limiting. When given adequate visibility into the sheet's contents, Neural Reasoning often proves to be a more advantageous approach, even for straightforward tasks that don't involve intricate calculations. The key advantage lies in its ability to perceive and interpret the finer details and context embedded within the data itself, which might be overlooked by a rigid symbolic approach. This allows the model to form a more holistic understanding of the problem, leading to superior results.
 "question": "What is the total landings (tonnes live weight) for Scotland in 2023, and how does it compare to the total landings for England, Wales, and N.I.? "
 "expected answer": "163802, 6987, much higher",

\begin{figure}[htbp]
    \centering
    \includegraphics[width=\columnwidth]{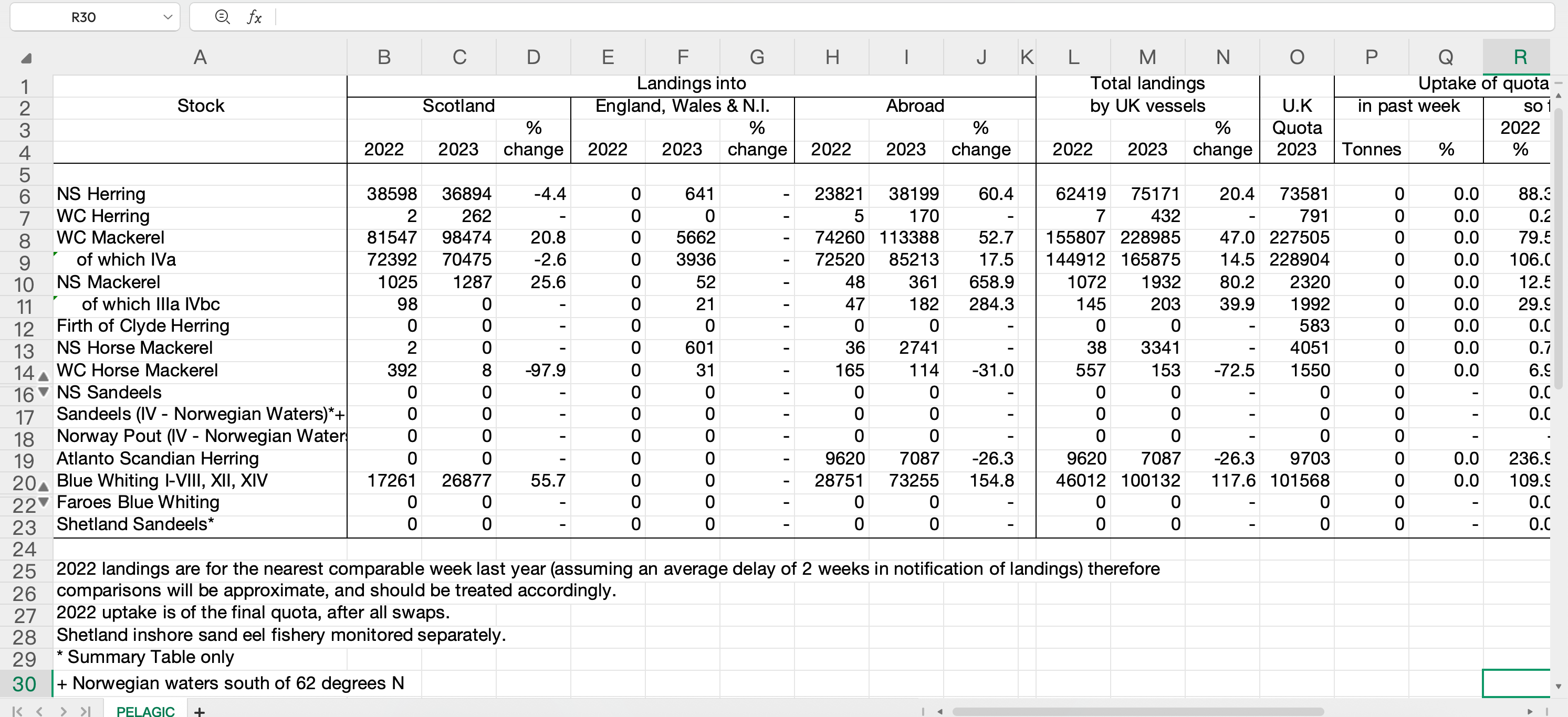}
    \caption{Case 4 Sheet}
    \label{fig:comparison}
\end{figure}

\subsubsection{Microsoft 365 copilot (analyst mode providing the excel file): Wrong}

\begin{Verbatim}
**Opening and inspecting file**I'm preparing to open the Excel file with Python and explore its sheets using openpyxl. The goal is to extract the total landings data for Scotland, England, Wales, and Northern Ireland in 2023.


**Checking Excel sheet content**I’ll explore the Excel sheet to find the total landings figures for 2023. Once I identify the relevant sheet and rows, I'll compare Scotland’s landings with those from England, Wales, and Northern Ireland.


```python
# Let's load the Excel file to see what sheets it contains
import pandas as pd
import numpy as np


# Load Excel file and inspect sheet names
xls = pd.ExcelFile('/mnt/data/business-table09.xlsx')
sheet_names = xls.sheet_names
sheet_names
```




```vbnet
Status: Success
Result: ['PELAGIC']
```


**Loading and examining data**Opening the "PELAGIC" sheet from the Excel file, extracting Scotland's rows, and checking the relevant columns, including identifying the 2023 column.


```python
# Load the Pelagic sheet to inspect its content properly
import pandas as pd


df = pd.read_excel('/mnt/data/business-table09.xlsx', sheet_name='PELAGIC')
print(df.head(20))
```




```vbnet
Status: Success
Stdout:                                Unnamed: 0 Landings into  Unnamed: 2  \
0                                   Stock      Scotland         NaN  
1                                     NaN           NaN         NaN  
2                                     NaN          2022     2023.00  
3                                     NaN           NaN         NaN  
4                              NS Herring       38597.9    36893.62  
5                              WC Herring          1.78      262.09  
6                             WC Mackerel      81546.81    98473.65  
7                            of which IVa      72391.79    70475.45  
8                             NS Mackerel       1024.67     1287.44  
9                      of which IIIa IVbc         97.74        0.00  
10                 Firth of Clyde Herring             0        0.00  
11                      NS Horse Mackerel           1.5        0.00  
12                      WC Horse Mackerel        392.27        8.36  
13                     NS Blue Whiting EU             0        0.00  
14                            NS Sandeels             0        0.00  
15     Sandeels (IV - Norwegian Waters)*+             0        0.00  
16  Norway Pout (IV - Norwegian Waters)*+             0        0.00  
17               Atlanto Scandian Herring             0        0.00  
18          Blue Whiting I-VIII, XII, XIV       17261.3    26877.05  
19              of which Bay of Biscay BW       17261.3    26877.05  


  Unnamed: 3             Unnamed: 4  Unnamed: 5 Unnamed: 6 Unnamed: 7  \
0         NaN  England, Wales & N.I.         NaN        NaN     Abroad  
1           %                    NaN         NaN          %        NaN  
2      change                   2022    2023.000     change       2022  
3         NaN                    NaN         NaN        NaN        NaN  
4   -4.415473                      0     640.896          -   23821.22  
5           -                      0       0.000          -       5.35  
6   20.757207                      0    5662.283          -   74260.35  
7   -2.647179                      0    3935.516          -   72519.96  
8   25.644354                      0      52.293          -       47.6  
9           -                      0      21.304          -      47.25  
10          -                      0       0.000          -          0  
11          -                      0     600.660          -      36.01  
12 -97.868815                      0      30.816          -     165.05  
13          -                      0       0.000          -       3.21  
14          -                      0       0.000          -          0  
15          -                      0       0.000          -          0  
16          -                      0       0.000          -          0  
17          -                      0       0.000          -    9620.33  
18  55.706986                      0       0.000          -   28750.62  
19  55.706986                      0       0.000          -          0  


    Unnamed: 8  Unnamed: 9  ...  Unnamed: 13 Unnamed: 14  Uptake of quota  \
0           NaN         NaN  ...          NaN         U.K     in past week  
1           NaN           %  ...            %       Quota              NaN  
2     2023.0000      change  ...       change        2023           Tonnes  
3           NaN         NaN  ...          NaN         NaN              NaN  
4    38198.5493   60.355134  ...    20.429534   73580.634                0  
5      170.2330           -  ...            -         791                0  
6   113387.9450   52.689753  ...    46.966931  227504.661                0  
7    85213.2170   17.503122  ...    14.466087  228904.161                0  
8      361.2340  658.894958  ...    80.185681    2320.289                0  
9      181.5830  284.302646  ...    39.931719    1991.789                0  
10       0.0000           -  ...            -     582.977                0  
11    2740.8040           -  ...            -    4050.982                0  
12     113.9670  -30.950015  .....
```


**Examining Excel file content**I am reviewing the Excel file, specifically the 'PELAGIC' sheet. It has data on landings for Scotland, England, Wales, and Northern Ireland, with values for 2022 and 2023.


**Processing Excel data**I am examining the structure of the 'PELAGIC' sheet in the Excel file, attempting to read it fully using pandas. I’ll check column names and rows to find the total landings for 2023.


**Reading sheet without headers**I am attempting to load the Excel sheet without using headers and will check the shape of the data to better understand its structure.


```python
# Load the excel data to inspect structure with a little less header structure.


df = pd.read_excel('/mnt/data/business-table09.xlsx', sheet_name='PELAGIC', header=None)
print(df.head(30))


```




```vbnet
Status: Success
Stdout:                                                    0              1   \
0                                                 NaN  Landings into  
1                                               Stock       Scotland  
2                                                 NaN            NaN  
3                                                 NaN           2022  
4                                                 NaN            NaN  
5                                          NS Herring        38597.9  
6                                          WC Herring           1.78  
7                                         WC Mackerel       81546.81  
8                                        of which IVa       72391.79  
9                                         NS Mackerel        1024.67  
10                                 of which IIIa IVbc          97.74  
11                             Firth of Clyde Herring              0  
12                                  NS Horse Mackerel            1.5  
13                                  WC Horse Mackerel         392.27  
14                                 NS Blue Whiting EU              0  
15                                        NS Sandeels              0  
16                 Sandeels (IV - Norwegian Waters)*+              0  
17              Norway Pout (IV - Norwegian Waters)*+              0  
18                           Atlanto Scandian Herring              0  
19                      Blue Whiting I-VIII, XII, XIV        17261.3  
20                          of which Bay of Biscay BW        17261.3  
21                                Faroes Blue Whiting              0  
22                                 Shetland Sandeels*              0  
23                                                NaN            NaN  
24  2022 landings are for the nearest comparable w...            NaN  
25  comparisons will be approximate, and should be...            NaN  
26  2022 uptake is of the final quota, after all s...            NaN  
27  Shetland inshore sand eel fishery monitored se...            NaN  
28                               * Summary Table only            NaN  
29           + Norwegian waters south of 62 degrees N            NaN  


         2          3                      4         5       6         7   \
0        NaN        NaN                    NaN       NaN     NaN       NaN  
1        NaN        NaN  England, Wales & N.I.       NaN     NaN    Abroad  
2        NaN          %                    NaN       NaN       %       NaN  
3    2023.00     change                   2022  2023.000  change      2022  
4        NaN        NaN                    NaN       NaN     NaN       NaN  
5   36893.62  -4.415473                      0   640.896       -  23821.22  
6     262.09          -                      0     0.000       -      5.35  
7   98473.65  20.757207                      0  5662.283       -  74260.35  
8   70475.45  -2.647179                      0  3935.516       -  72519.96  
9    1287.44  25.644354                      0    52.293       -      47.6  
10      0.00          -                      0    21.304       -     47.25  
11      0.00          -                      0     0.000       -         0  
12      0.00          -                      0   600.660       -     36.01  
13      8.36 -97.868815                      0    30.816       -    165.05  
14      0.00          -                      0     0.000       -      3.21  
15      0.00          -                      0     0.000       -         0  
16      0.00          -                      0     0.000       -         0  
17      0.00          -                      0     0.000       -         0  
18      0.00          -                      0     0.000       -   9620.33  
19  26877.05  55.706986                      0     0.000       -  28750.62  
20  26877.05  55.706986                      0     0.000       -         0  
21      0.00          -                      0     0.000       -         0  
22      0.00          -                      0     0.000       - ...
```


**Interpreting Excel file data**I’m reviewing an Excel sheet titled "PELAGIC" to gather landing data by region for 2023. I'll focus on columns relating to "Scotland" and "England, Wales & N.I." and ensure the table is read correctly.


**Analyzing total landings data**I am identifying the total landings for Scotland in 2023 by reviewing the relevant Excel table. I'll focus on the "Total" rows for each region and ensure proper parsing of headers and fish species data.


```python
# Let's load the entire sheet as dataframe in a single multiindex
import pandas as pd


# Load the sheet as is with header=0, but the table likely has a merged header
xls = pd.read_excel('/mnt/data/business-table09.xlsx', sheet_name='PELAGIC', header=0)
print(xls.head(50))
```




```vbnet
Status: Success
Stdout:                                            Unnamed: 0 Landings into  \
0                                               Stock      Scotland  
1                                                 NaN           NaN  
2                                                 NaN          2022  
3                                                 NaN           NaN  
4                                          NS Herring       38597.9  
5                                          WC Herring          1.78  
6                                         WC Mackerel      81546.81  
7                                        of which IVa      72391.79  
8                                         NS Mackerel       1024.67  
9                                  of which IIIa IVbc         97.74  
10                             Firth of Clyde Herring             0  
11                                  NS Horse Mackerel           1.5  
12                                  WC Horse Mackerel        392.27  
13                                 NS Blue Whiting EU             0  
14                                        NS Sandeels             0  
15                 Sandeels (IV - Norwegian Waters)*+             0  
16              Norway Pout (IV - Norwegian Waters)*+             0  
17                           Atlanto Scandian Herring             0  
18                      Blue Whiting I-VIII, XII, XIV       17261.3  
19                          of which Bay of Biscay BW       17261.3  
20                                Faroes Blue Whiting             0  
21                                 Shetland Sandeels*             0  
22                                                NaN           NaN  
23  2022 landings are for the nearest comparable w...           NaN  
24  comparisons will be approximate, and should be...           NaN  
25  2022 uptake is of the final quota, after all s...           NaN  
26  Shetland inshore sand eel fishery monitored se...           NaN  
27                               * Summary Table only           NaN  
28           + Norwegian waters south of 62 degrees N           NaN  


   Unnamed: 2 Unnamed: 3             Unnamed: 4  Unnamed: 5 Unnamed: 6  \
0          NaN        NaN  England, Wales & N.I.         NaN        NaN  
1          NaN          %                    NaN         NaN          %  
2      2023.00     change                   2022    2023.000     change  
3          NaN        NaN                    NaN         NaN        NaN  
4     36893.62  -4.415473                      0     640.896          -  
5       262.09          -                      0       0.000          -  
6     98473.65  20.757207                      0    5662.283          -  
7     70475.45  -2.647179                      0    3935.516          -  
8      1287.44  25.644354                      0      52.293          -  
9         0.00          -                      0      21.304          -  
10        0.00          -                      0       0.000          -  
11        0.00          -                      0     600.660          -  
12        8.36 -97.868815                      0      30.816          -  
13        0.00          -                      0       0.000          -  
14        0.00          -                      0       0.000          -  
15        0.00          -                      0       0.000          -  
16        0.00          -                      0       0.000          -  
17        0.00          -                      0       0.000          -  
18    26877.05  55.706986                      0       0.000          -  
19    26877.05  55.706986                      0       0.000          -  
20        0.00          -                      0       0.000          -  
21        0.00          -                      0       0.000          -  
22         NaN        NaN                    NaN         NaN        NaN  
23         NaN        NaN                    NaN         NaN        NaN  
24         NaN        NaN                    NaN         NaN        NaN  
25         ...
```


```python
# Let's extract data from the Excel file. We need the total landings for Scotland for the year 2023 and total landings for England, Wales, and N.I.


import pandas as pd


# Read data from the Excel sheet with no header because headers not clear.
xls = pd.read_excel('/mnt/data/business-table09.xlsx', sheet_name='PELAGIC', header=None)


# Inspect the general shape and possibly second set of rows are for indexes
#print(xls.iloc[:, :10]) as we want print first 30 rows and all columns.


# The question is total landings for Scotland in 2023 which are measured in 'tonnes live weight' so we need to look at table...
# Let's examine where generic totals might be: may be row with column 'Total' or similar.
# Let's check the row with 'Total' in different columns.
print(xls[xls.apply(lambda row: row.astype(str).str.contains('total', case=False, na=False).any(), axis=1)])


```




```vbnet
Status: Success
Stdout:     0              1   2    3    4   5    6    7   8    9   ...   13   14  \
0  NaN  Landings into NaN  NaN  NaN NaN  NaN  NaN NaN  NaN  ...  NaN  NaN  


               15   16   17   18  19  20  21  22 
0  Uptake of quota  NaN  NaN  NaN NaN NaN NaN     


[1 rows x 23 columns]


```


**Inspecting the data file**Looking at the Excel file structure to identify total landings tonnage for Scotland in 2023, then comparing it to the respective totals for England, Wales, and Northern Ireland.


**Analyzing the table structure**Manually examining the excel file to find landings data for Scotland and for England, Wales, and Northern Ireland in 2023, checking for a "Total" row or similar indicators.


**Inspecting the headers and rows**I’ll try to adjust the header row configuration by specifying multiple rows and review the first 30 rows in the dataframe to better understand the structure of the data and identify the total landings for Scotland and other regions.


```python
# Let's check the rows 0 to 35 as a matrix to see what content we have:
for i in range(35):
   print(i, df.iloc[i].values.tolist())
```




```vbnet
Status: Failure
Stdout: 0 [nan, 'Landings into', nan, nan, nan, nan, nan, nan, nan, nan, nan, 'Total landings', nan, nan, nan, 'Uptake of quota', nan, nan, nan, nan, nan, nan, ' ']
1 ['Stock', 'Scotland', nan, nan, 'England, Wales & N.I.', nan, nan, 'Abroad', nan, nan, nan, 'by UK vessels', nan, nan, 'U.K', 'in past week', nan, 'so far in:', nan, nan, nan, nan, 'U.K']
2 [nan, nan, nan, '%', nan, nan, '%', nan, nan, '%', nan, nan, nan, '%', 'Quota', nan, nan, 2022, 2023, nan, nan, nan, 'Quota']
3 [nan, 2022, 2023.0, 'change', 2022, 2023.0, 'change', 2022, 2023.0, 'change', nan, 2022, 2023.0, 'change', 2023, 'Tonnes', '%', '%', '%', nan, nan, nan, 2022]
4 [nan, nan, nan, nan, nan, nan, nan, nan, nan, nan, nan, nan, nan, nan, nan, nan, nan, nan, nan, nan, nan, nan, nan]
5 ['NS Herring', 38597.9, 36893.62, -4.41547338067615, 0, 640.896, '-', 23821.22, 38198.5493, 60.3551342038737, nan, 62419.12, 75171.0553, 20.4295339312698, 73580.634, 0, 0, 88.274313255193, 102.161467241503, nan, nan, nan, 70710.4]
6 ['WC Herring', 1.78, 262.09, '-', 0, 0.0, '-', 5.35, 170.233, '-', nan, 7.13, 432.323, '-', 791, 0, 0, 0.175697986742564, 54.6552465233881, nan, nan, nan, 4058.1]
7 ['WC Mackerel', 81546.81, 98473.65, 20.757206811646, 0, 5662.28299999999, '-', 74260.35, 113387.945, 52.6897530108597, nan, 155807.16, 228985.001, 46.9669307880331, 227504.661, 0, 0, 79.518846325408, 100.65068556991, nan, nan, nan, 195937.4]
8 ['    of which IVa', 72391.79, 70475.45, -2.64717863724597, 0, 3935.516, '-', 72519.96, 85213.217, 17.5031218991296, nan, 144911.75, 165874.81, 14.4660871185394, 228904.161, 0, 0, 106.019774120835, 72.464742132844, nan, nan, nan, 136683.7]
9 ['NS Mackerel', 1024.67, 1287.44, 25.6443537919526, 0, 52.293, '-', 47.6, 361.234, 658.894957983193, nan, 1072.27, 1932.077, 80.1856808453095, 2320.289, 0, 0, 12.5426365656802, 83.2688083251699, nan, nan, nan, 8549]
10 ['     of which IIIa IVbc', 97.74, 0.0, '-', 0, 21.304, '-', 47.25, 181.583, 284.302645502646, nan, 144.99, 202.887, 39.9317194289262, 1991.789, 0, 0, 29.9380549246335, 10.1861693181356, nan, nan, nan, 484.3]
11 ['Firth of Clyde Herring', 0, 0.0, '-', 0, 0.0, '-', 0, 0.0, '-', nan, 0, 0.0, '-', 582.977, 0, 0, 0, 0, nan, nan, nan, 641.3]
12 ['NS Horse Mackerel', 1.5, 0.0, '-', 0, 600.66, '-', 36.01, 2740.804, '-', nan, 37.51, 3341.464, '-', 4050.982, 0, 0, 0.656297022080694, 82.4852838151342, nan, nan, nan, 5715.4]
13 ['WC Horse Mackerel', 392.27, 8.36, -97.8688148469167, 0, 30.816, '-', 165.05, 113.967, -30.9500151469252, nan, 557.32, 153.143, -72.5215316155889, 1550.175, 0, 0, 6.89991581241024, 9.87907816859387, nan, nan, nan, 8077.2]
14 ['NS Blue Whiting EU', 0, 0.0, '-', 0, 0.0, '-', 3.21, 0.0, '-', nan, 3.21, 0.0, '-', 0, 0, '-', nan, '-', nan, nan, nan, nan]
15 ['NS Sandeels', 0, 0.0, '-', 0, 0.0, '-', 0, 0.0, '-', nan, 0, 0.0, '-', 0, 0, '-', 0, '-', nan, nan, nan, 1741.2]
16 ['Sandeels (IV - Norwegian Waters)*+', 0, 0.0, '-', 0, 0.0, '-', 0, 0.0, '-', nan, 0, 0.0, '-', 0, 0, '-', 0, '-', nan, nan, nan, 0]
17 ['Norway Pout (IV - Norwegian Waters)*+', 0, 0.0, '-', 0, 0.0, '-', 0, 0.141, '-', nan, 0, 0.141, '-', 0, 0, '-', '-', '-', nan, nan, nan, 0]
18 ['Atlanto Scandian Herring', 0, 0.0, '-', 0, 0.0, '-', 9620.33, 7087.29, -26.3300739163833, nan, 9620.33, 7087.29, -26.3300739163833, 9703, 0, 0, 236.878093221382, 73.0422549726889, nan, nan, nan, 4061.3]
19 ['Blue Whiting I-VIII, XII, XIV', 17261.3, 26877.05, 55.7069861482044, 0, 0.0, '-', 28750.62, 73254.908, 154.794185308004, nan, 46011.92, 100131.958, 117.621777139489, 101567.503, 0, 0, 109.897320394286, 98.586609931722, nan, nan, nan, 41868.1]
20 ['   of which Bay of Biscay BW', 17261.3, 26877.05, 55.7069861482044, 0, 0.0, '-', 0, 0.0, '-', nan, 17261.3, 26877.05, 55.7069861482044, 0, 0, '-', nan, '-', nan, nan, nan, nan]
21 ['Faroes Blue Whiting', 0, 0.0, '-', 0, 0.0, '-', 0, 0.0, '-', nan, 0, 0.0, '-', 0, 0, '-', 0, '-', nan, nan, nan, 713]
22 ['Shetland Sandeels*', 0, 0.0, '-', 0, 0.0, '-', 0, 0.0, '-', nan, 0, 0.0, '-', 0, 0, '-', 0, '-', nan, nan, nan, 0]
23 [nan, nan, nan, nan, nan, nan, nan, nan, nan, nan, nan, nan, nan, nan, nan...
__ERR__Stderr: single positional indexer is out-of-bounds---------------------------------------------------------------------------
IndexError                                Traceback (most recent call last)
Cell In[14], line 6
     4 # Let's check the rows 0 to 35 as a matrix to see what content we have:
     5 for i in range(35):
----> 6     print(i, df.iloc[i].values.tolist())


File ~/snenv/lib/python3.12/site-packages/pandas/core/indexing.py:1191, in _LocationIndexer.__getitem__(self, key)
  1189 maybe_callable = com.apply_if_callable(key, self.obj)
  1190 maybe_callable = self._check_deprecated_callable_usage(key, maybe_callable)
-> 1191 return self._getitem_axis(maybe_callable, axis=axis)


File ~/snenv/lib/python3.12/site-packages/pandas/core/indexing.py:1752, in _iLocIndexer._getitem_axis(self, key, axis)
  1749     raise TypeError("Cannot index by location index with a non-integer key")
  1751 # validate the location
-> 1752 self._validate_integer(key, axis)
  1754 return self.obj._ixs(key, axis=axis)


File ~/snenv/lib/python3.12/site-packages/pandas/core/indexing.py:1685, in _iLocIndexer._validate_integer(self, key, axis)
  1683 len_axis = len(self.obj._get_axis(axis))
  1684 if key >= len_axis or key < -len_axis:
-> 1685     raise IndexError("single positional indexer is out-of-bounds")


IndexError: single positional indexer is out-of-bounds
__ERR__
```


**Analyzing file content**I am exploring the "business-table09.xlsx" file, specifically targeting the "PELAGIC" sheet. I need to find the total landings data for each region (Scotland, England, Wales, N.I.) in 2023.


**Parsing data structure**I am working on structuring the Excel sheet by mapping columns and rows to identify the "Total landings" row for each region in 2023, and confirming the shape of the data.


**Refining header structure**I am experimenting with different header configurations, including a multi-index approach, to better parse and structure the data in the Excel sheet, specifically focusing on the “Total landings” row for 2023.


```python
# Try reading with a multi-index header, let's check how many header rows there are.


import pandas as pd


df_multi = pd.read_excel('/mnt/data/business-table09.xlsx', sheet_name='PELAGIC', header=[0,1,2,3])
print(df_multi.head(10))
```




```vbnet
Status: Success
Stdout:         Unnamed: 0_level_0      Landings into                                \
                    Stock           Scotland                                
       Unnamed: 0_level_2 Unnamed: 1_level_2 Unnamed: 2_level_2          %  
       Unnamed: 0_level_3               2022               2023     change  
0                      NaN                NaN                NaN        NaN  
1               NS Herring           38597.90           36893.62  -4.415473  
2               WC Herring               1.78             262.09          -  
3              WC Mackerel           81546.81           98473.65  20.757207  
4             of which IVa           72391.79           70475.45  -2.647179  
5              NS Mackerel            1024.67            1287.44  25.644354  
6       of which IIIa IVbc              97.74               0.00          -  
7   Firth of Clyde Herring               0.00               0.00          -  
8        NS Horse Mackerel               1.50               0.00          -  
9        WC Horse Mackerel             392.27               8.36 -97.868815  


                                               \
 England, Wales & N.I.                                       Abroad  
    Unnamed: 4_level_2 Unnamed: 5_level_2      % Unnamed: 7_level_2  
                  2022               2023 change               2022  
0                   NaN                NaN    NaN                NaN  
1                   0.0            640.896      -           23821.22  
2                   0.0              0.000      -               5.35  
3                   0.0           5662.283      -           74260.35  
4                   0.0           3935.516      -           72519.96  
5                   0.0             52.293      -              47.60  
6                   0.0             21.304      -              47.25  
7                   0.0              0.000      -               0.00  
8                   0.0            600.660      -              36.01  
9                   0.0             30.816      -             165.05  


                                 ... Total landings              \
                                 ...  by UK vessels         U.K  
 Unnamed: 8_level_2           %  ...              %       Quota  
               2023      change  ...         change        2023  
0                NaN         NaN  ...            NaN         NaN  
1         38198.5493   60.355134  ...      20.429534   73580.634  
2           170.2330           -  ...              -     791.000  
3        113387.9450   52.689753  ...      46.966931  227504.661  
4         85213.2170   17.503122  ...      14.466087  228904.161  
5           361.2340  658.894958  ...      80.185681    2320.289  
6           181.5830  284.302646  ...      39.931719    1991.789  
7             0.0000           -  ...              -     582.977  
8          2740.8040           -  ...              -    4050.982  
9           113.9670  -30.950015  ...     -72.521532    1550.175  


     Uptake of quota                                                          \
        in past week                      so far in:                          
 Unnamed: 15_level_2 Unnamed: 16_level_2        2022        2023              
              Tonnes                   %           %           % %.1 %.2 %.3  
0                 NaN                 NaN         NaN         NaN NaN NaN NaN  
1                 0.0                   0   88.274313  102.161467 NaN NaN NaN  
2                 0.0                   0    0.175698   54.655247 NaN NaN NaN  
3                 0.0                   0   79.518846  100.650686 NaN NaN NaN  
4                 0.0                   0  106.019774   72.464742 NaN NaN NaN  
5                 0.0                   0   12.542637   83.268808 NaN NaN NaN  
6                 0.0                   0   29.938055   10.186169 NaN NaN NaN  
7                 0.0                   0           0           0 NaN NaN NaN  
8                 0.0                   0    0.6562...
```


**Examining data structure**I’m reviewing an Excel file to extract the total landings for Scotland in 2023 and compare it with those of England, Wales, and Northern Ireland. The table has a multi-index format.


**Checking column structure**I’m reassessing how to handle the multi-index header rows in the Excel file, particularly to find the correct columns for "Total landings" and adjust the selection accordingly.


```python
# Let's check column names of df_multi.
print(df_multi.columns.tolist())
```




```vbnet
Status: Success
Stdout: [('Unnamed: 0_level_0', 'Stock', 'Unnamed: 0_level_2', 'Unnamed: 0_level_3'), ('Landings into', 'Scotland', 'Unnamed: 1_level_2', 2022), ('Landings into', 'Scotland', 'Unnamed: 2_level_2', 2023), ('Landings into', 'Scotland', '%', 'change'), ('Landings into', 'England, Wales & N.I.', 'Unnamed: 4_level_2', 2022), ('Landings into', 'England, Wales & N.I.', 'Unnamed: 5_level_2', 2023), ('Landings into', 'England, Wales & N.I.', '%', 'change'), ('Landings into', 'Abroad', 'Unnamed: 7_level_2', 2022), ('Landings into', 'Abroad', 'Unnamed: 8_level_2', 2023), ('Landings into', 'Abroad', '%', 'change'), ('Landings into', 'Abroad', '%', 'change.1'), ('Total landings', 'by UK vessels', 'Unnamed: 11_level_2', 2022), ('Total landings', 'by UK vessels', 'Unnamed: 12_level_2', 2023), ('Total landings', 'by UK vessels', '%', 'change'), ('Total landings', 'U.K', 'Quota', 2023), ('Uptake of quota', 'in past week', 'Unnamed: 15_level_2', 'Tonnes'), ('Uptake of quota', 'in past week', 'Unnamed: 16_level_2', '%'), ('Uptake of quota', 'so far in:', 2022, '%'), ('Uptake of quota', 'so far in:', 2023, '%'), ('Uptake of quota', 'so far in:', 2023, '%.1'), ('Uptake of quota', 'so far in:', 2023, '%.2'), ('Uptake of quota', 'so far in:', 2023, '%.3'), (' ', 'U.K', 'Quota', 2022)]


```


**Analyzing the multi-index DataFrame**Exploring the multi-index columns of the DataFrame and identifying the relevant data for calculating the total landings for Scotland and "England, Wales & N.I." in 2023.


**Identifying total landings row**Exploring the structure of the DataFrame to find the "Total landings" row and figure out the correct columns to sum in order to get the total landings for Scotland in 2023.


**Checking for total row**I am examining the DataFrame to look for "Total" among the stock labels to identify the appropriate row that might give the total landings for Scotland in 2023.


```python
# Print first column as labels
labels = df_multi[('Unnamed: 0_level_0', 'Stock', 'Unnamed: 0_level_2', 'Unnamed: 0_level_3')]
print(labels.tolist()[:30])
```




```vbnet
Status: Success
Stdout: [nan, 'NS Herring', 'WC Herring', 'WC Mackerel', '    of which IVa', 'NS Mackerel', '     of which IIIa IVbc', 'Firth of Clyde Herring', 'NS Horse Mackerel', 'WC Horse Mackerel', 'NS Blue Whiting EU', 'NS Sandeels', 'Sandeels (IV - Norwegian Waters)*+', 'Norway Pout (IV - Norwegian Waters)*+', 'Atlanto Scandian Herring', 'Blue Whiting I-VIII, XII, XIV', '   of which Bay of Biscay BW', 'Faroes Blue Whiting', 'Shetland Sandeels*', nan, '2022 landings are for the nearest comparable week last year (assuming an average delay of 2 weeks in notification of landings) therefore', 'comparisons will be approximate, and should be treated accordingly.', '2022 uptake is of the final quota, after all swaps.', 'Shetland inshore sand eel fishery monitored separately.', '* Summary Table only', '+ Norwegian waters south of 62 degrees N']


```


**Analyzing Excel data extraction**Exploring how to aggregate the total landings for Scotland in 2023 by summing the species data by region, while also considering totals for the UK and comparison with other regions.


**Confirming data sources**The "Total landings" row likely contains the overall total for landings by UK vessels in 2023. Data for Scotland in 2023 appears to be in specific columns with regions like "Landings into, Scotland".


**Summing values for regions**The goal is to sum the landings for each species by region (Scotland and England, Wales & N.I.) for 2023, while also verifying totals in the "Total landings by UK vessels" columns to ensure accuracy.


**Computing region totals**I will sum the landings for Scotland and England, Wales & N.I. in 2023 by filtering out rows with non-numeric labels or commentary, and then calculating the totals for each region.


```python
# Let's sum up the Scotland landings for 2023 and England, Wales, N.I. for 2023.
# Use the multi-index df_multi we read.


col_scot_2023 = ('Landings into', 'Scotland', 'Unnamed: 2_level_2', 2023)
col_eng_2023 = ('Landings into', 'England, Wales & N.I.', 'Unnamed: 5_level_2', 2023)


# Sum these columns across all rows.
scot_sum = pd.to_numeric(df_multi[col_scot_2023], errors='coerce').sum()
eng_sum = pd.to_numeric(df_multi[col_eng_2023], errors='coerce').sum()


scot_sum, eng_sum


```




```vbnet
Status: Success
Result: (261154.71, 10943.76799999999)
```


**Clarifying landing numbers**I'm reviewing the total landings for Scotland and England, Wales & N.I. for 2023. It looks like the sums might need adjustments to ensure we're correctly interpreting species and stock data.


**Verifying data structure**I’m examining the structure of the table to narrow down the relevant fish species rows while filtering out non-numeric values. It’s important to determine if 261154.71 is the actual total for Scotland.


**Summing total landings**I’m reevaluating the approach to summing landings for both Scotland and England, Wales & N.I. in 2023. The goal is to compare total landings while considering exclusion for flags like species or percentages.


**Refining row inclusions**I’m ensuring the sums account for all relevant species rows and excluding non-numeric or NaN rows. Next, I’ll compare the sums for Scotland and England, Wales & N.I., while checking the individual species values.


```python
# Get detailed data of landings for each species for Scotland and E/W/NI for 2023
species = df_multi[('Unnamed: 0_level_0', 'Stock', 'Unnamed: 0_level_2', 'Unnamed: 0_level_3')]
scot_2023 = df_multi[('Landings into', 'Scotland', 'Unnamed: 2_level_2', 2023)]
eng_2023 = df_multi[('Landings into', 'England, Wales & N.I.', 'Unnamed: 5_level_2', 2023)]
data = pd.DataFrame({'Species': species, 'Scot_2023': scot_2023, 'ENG_2023': eng_2023})
print(data.head(20))
```




```vbnet
Status: Success
Stdout:                                   Species  Scot_2023  ENG_2023
0                                     NaN        NaN       NaN
1                              NS Herring   36893.62   640.896
2                              WC Herring     262.09     0.000
3                             WC Mackerel   98473.65  5662.283
4                            of which IVa   70475.45  3935.516
5                             NS Mackerel    1287.44    52.293
6                      of which IIIa IVbc       0.00    21.304
7                  Firth of Clyde Herring       0.00     0.000
8                       NS Horse Mackerel       0.00   600.660
9                       WC Horse Mackerel       8.36    30.816
10                     NS Blue Whiting EU       0.00     0.000
11                            NS Sandeels       0.00     0.000
12     Sandeels (IV - Norwegian Waters)*+       0.00     0.000
13  Norway Pout (IV - Norwegian Waters)*+       0.00     0.000
14               Atlanto Scandian Herring       0.00     0.000
15          Blue Whiting I-VIII, XII, XIV   26877.05     0.000
16              of which Bay of Biscay BW   26877.05     0.000
17                    Faroes Blue Whiting       0.00     0.000
18                     Shetland Sandeels*       0.00     0.000
19                                    NaN        NaN       NaN


```


**Comparing landings data**Calculating total landings for Scotland and England/Wales/N.I. in 2023, I find Scotland's landings to be approximately 23.87 times higher than for England, Wales, and Northern Ireland.


**Verifying landings data**Looking into how "Total landings" are structured, I confirm that Scotland's landings in 2023 are approximately 23.87 times greater than those for England, Wales, and Northern Ireland combined.


**Finalizing answer**Confirming Scotland's total landings in 2023 as 261,155 tonnes, compared to 10,944 tonnes for England, Wales, and N.I. Scotland's landings are approximately 23.9 times higher.


In **2023**, the total landings (tonnes live weight) were:


- **Scotland**: **261,155 tonnes**
- **England, Wales & Northern Ireland**: **10,944 tonnes**


### Comparison:
Scotland's landings were approximately **24 times greater** than those of England, Wales, and Northern Ireland combined.


Would you like a breakdown by species or a visual comparison chart?

\end{Verbatim}

\subsubsection{Microsoft 365 copilot (analyst mode providing the sheet content serialization) : Correct}

\begin{Verbatim}
    

**Parsing table data**I'm looking at a table labeled "PELAGIC" and working on extracting the total landings for Scotland, England, Wales, and Northern Ireland in 2023. I'll parse the CSV-like content for the relevant data.


**Clarifying table structure**The "PELAGIC" table is organized with headings for regions like Scotland and England, Wales & N.I. I will sum the 2023 landings for these regions based on the relevant columns.


**Verifying table structure**I'm clarifying the table layout, focusing on column C for Scotland's 2023 landings and column F for England, Wales & N.I., then I’ll sum the values for comparison.


**Preparing for data parsing**I'm analyzing the structure of the "PELAGIC" sheet, focusing on summing columns C (Scotland 2023) and F (England, Wales & N.I. 2023) to answer the total landings question. I'll parse the rows and calculate the sums accordingly.


**Summing landings by region**I am preparing to sum the 2023 landings for Scotland (column C) and England, Wales & N.I. (column F) across rows 6 to 24, excluding subcategories like "of which." I'll calculate and compare the totals.


**Finalizing the sum calculation**I’m proceeding with summing the 2023 landings for Scotland (Column C) and England, Wales & N.I. (Column F) by species from rows 6 to 23, excluding breakdown rows like "of which."


**Summing final landings values**I’ve reviewed the data and will now sum the relevant 2023 landings for Scotland (column C) and England, Wales & N.I. (column F), excluding breakdown rows like "of which," for final totals. I'll proceed with the Python code implementation.


**Summing values from table rows**Filtering and summing values for Scotland and England/Wales/N.I. from specific rows, excluding those with "of which" in column A, to get precise totals for 2023.


**Computing regional sums**Parsing numeric values from columns C and F to compute sums for Scotland and England, Wales & NI, filtering out subcategory rows and excluding entries starting with "of which".


**Implementing summing logic**Writing code to sum column values based on specific row conditions, excluding rows containing "of which" and extracting relevant data for Scotland and England, Wales & NI from specified columns.


**Finalizing the answer**I am concluding that Scotland had 163,802.21 tonnes of landings in 2023, while England, Wales & N.I. had just 6,986.95 tonnes. The difference is significant.


**Computing ratio difference**Scotland's landings in 2023 were about 23.5 times greater than that of England, Wales & N.I., with totals of 163,802.21 tonnes versus 6,986.95 tonnes.


In 2023:


- **Scotland** landed approximately **163,802 tonnes**.
- **England, Wales & Northern Ireland** landed approximately **6,987 tonnes**.


**Comparison**: Scotland's landings were **over 23 times greater** than those of England, Wales & N.I.


Would you like a visual comparison (e.g., a bar chart) or a breakdown by species?

\end{Verbatim}

\section{\huge Experimental Settings and Datasets}

Our experiments evaluate our method on three public distinct datasets: \textbf{MultiHiertt}, \textbf{SpreadsheetBench}, and \textbf{RealHitBench}.

For \textbf{MultiHiertt}, we used the first 300 QA pairs from the development set. For \textbf{SpreadsheetBench}, we use official 200 data samples, resulting in a total of 600 question-output pairs. This includes 369 pairs categorized as ``Cell-Level Manipulation'' and 231 pairs for ``Sheet-Level Manipulation''. For \textbf{RealHitBench}, we selected the first 300 question-answer pairs from the \texttt{QA\_final} set, excluding visualization tasks. This resulted in 131 pairs for ``Fact Checking'', 78 for ``Data Analysis'', and 91 for ``Numerical Reasoning''.

We used \textbf{GPT-4.1 as the backbone model} to test several agents, including StructGPT, SheetAgent, and our proposed \textbf{SheetBrain}. For our SheetBrain tests, we implemented a 10k token limit for the sheet content within the agent's internal context. The API calls were made with a \texttt{temperature} of 0.

To ensure robust evaluation, we adopted an \textbf{LLM-as-a-judge} methodology, using a GPT-4.1 backbone for automatic evaluation. The evaluation prompt template is provided here (MultiHiertt as example).

\begin{Verbatim}
evaluation_prompt = f"""
You are an expert evaluator for spreadsheet Q&A tasks, specifically evaluating MultiHiertt results.

QUESTION: {question}

EXPECTED ANSWER: {expected_answer}

AGENT ANSWER: {agent_answer}

TAGS: {tags_str}

Your task is to evaluate whether the agent's answer is fundamentally correct compared to the expected answer. Consider:

1. **Numerical Accuracy**: For numerical questions, check if the agent's calculations and final numbers match the expected values (allow for minor rounding differences like 0.1% variance).

2. **Factual Accuracy**: For factual questions, verify if the agent identifies the correct information, even if expressed differently.

3. **Year/Date Matching**: For questions asking for specific years or dates, the answer must be exactly correct.

4. **Value Matching**: For questions asking for specific values, names, or categories, check for exact or semantically equivalent matches.

5. **Financial/Business Data**: For questions involving financial metrics, ratios, percentages, ensure numerical accuracy.

6. **Comparison Questions**: For questions asking about comparisons (greater/less, highest/lowest), verify the logical correctness.

7. **Yes/No Questions**: For boolean questions, ensure the answer matches the expected true/false or yes/no response.

8. **Semantic Equivalence**: Answers may be phrased differently but convey the same meaning (e.g., "2006" vs "The year 2006" vs "2006 had the highest value").

**CRITICAL EVALUATION RULE - Data Availability Claims:**
- If the expected answer is a specific numerical value, year, percentage, ratio, or any concrete data point, this indicates that the required data IS PRESENT in the spreadsheet.
- If the agent claims "no data available", "cannot be determined", "data not present", etc., but the expected answer is a specific value, the agent's answer is INCORRECT.
- The presence of an expected answer implies the data exists and should be found by a competent agent.
- Only mark "no data available" claims as correct if the expected answer is also indicating missing data (e.g., "N/A", "Not available", etc.).


**Special Considerations for MultiHiertt:**
- Focus on the final factual answer rather than the reasoning process
- Allow for different phrasings of the same correct information
- Be strict about numerical precision but allow minor formatting differences
- Consider context-appropriate answers (e.g., currency formatting, percentage notation)
- For ratio questions, accept both decimal and percentage formats if they represent the same value
- For year questions, accept answers that clearly identify the correct year

Please provide your evaluation in the following JSON format:
{{
    "is_correct": true/false,
    "confidence_score": 0.0-1.0,
    "reasoning": "Brief explanation of your evaluation decision"
}}

Be fair but precise in your evaluation. If the core factual answer is correct despite minor presentation differences, mark it as correct.
"""
\end{Verbatim}

Our \textbf{SheetBench} dataset was compiled from 69 case data selected from 11 public datasets, including HiTab, MiMoTable, RealHitBench, MultiHiertt, DABStep, SheetCopilot, SpreadsheetBench, SheetAgent, NSF, MMQA, CondaBench. These cases are categorized into Complex Tables, Multi-table Layouts, Large Sheets, and Editing Operations. We manually reviewed and calibrated some queries and labels to correct errors present in the original data.

Additionally, results for BizChat Analyst and ChatGPT (4o) were obtained directly from their respective online interfaces by uploading the Excel files and corresponding queries.

\subsection{Different Serialization Examples}
\begin{itemize}
    \item Pure Markdown
\end{itemize}
\begin{verbatim}
| Time period |  |  | 2023-Q3 | 2023-Q4 | 2024-Q1 | 2024-Q2 | 2024-Q3 |      |
|             | Combined transaction | Combined unit of measure |  |  |  |  |  |  |      |
|             | Adjustment: Calendar and seasonally adjusted |  |  |  |  |  |  |


\end{verbatim}

\begin{itemize}
    \item Markdown with Cell Pos.
\end{itemize}
\begin{Verbatim}
| B6: Time period [MERGED 1x3] | C6: ~ | D6: ~ | E6: 2023-Q3 | F6: 2023-Q4 | G6: 2024-Q1 | H6: 2024-Q2 | I6: 2024-Q3 | J6:     |
| A7:                          | B7: Combined transaction | C7: Combined unit of measure | D7: | E7: | F7: | G7: | H7: | I7: | J7: |
| A8:                          | B8: Adjustment: Calendar and seasonally adjusted [MERGED 1x7] | C8: ~ | D8: ~ | E8: ~ | F8: ~ | G8: ~ | H8: ~ |


\end{Verbatim}

\begin{itemize}
    \item Pure HTML
\end{itemize}
\begin{verbatim}
<tr>
  <td>Time period</td>
  <td></td>
  <td></td>
  <td>2023-Q3</td>
  <td>2023-Q4</td>
  <td>2024-Q1</td>
  <td>2024-Q2</td>
  <td>2024-Q3</td>
  <td></td>
</tr>
<tr>
  <td></td>
  <td>Combined transaction</td>
  <td>Combined unit of measure</td>
  <td></td>
  <td></td>
  <td></td>
  <td></td>
  <td></td>
  <td></td>
</tr>
<tr>
  <td></td>
  <td>Adjustment: Calendar and seasonally adjusted</td>
  <td></td>
  <td></td>
  <td></td>
  <td></td>
  <td></td>
  <td></td>
</tr>

\end{verbatim}

\begin{itemize}
    \item MD-like HTML + CellPos.
\end{itemize}
\begin{verbatim}
<tr r>
  <td c>B6: Time period [MERGED 1x3]</td>
  <td>C6: ~</td>
  <td>D6: ~</td>
  <td>E6: 2023-Q3</td>
  <td>F6: 2023-Q4</td>
  <td>G6: 2024-Q1</td>
  <td>H6: 2024-Q2</td>
  <td>I6: 2024-Q3</td>
  <td>J6:</td>
</tr>
<tr>
  <td>A7:</td>
  <td>B7: Combined transaction</td>
  <td>C7: Combined unit of measure</td>
  <td>D7:</td>
  <td>E7:</td>
  <td>F7:</td>
  <td>G7:</td>
  <td>H7:</td>
  <td>I7:</td>
  <td>J7:</td>
</tr>
<tr>
  <td>A8:</td>
  <td>B8: Adjustment: Calendar and seasonally adjusted [MERGED 1x7]</td>
  <td>C8: ~</td>
  <td>D8: ~</td>
  <td>E8: ~</td>
  <td>F8: ~</td>
  <td>G8: ~</td>
  <td>H8: ~</td>
</tr>

\end{verbatim}

\begin{itemize}
    \item HTML + Colspan + CellPos.
\end{itemize}
\begin{verbatim}
<tr>
  <td colspan="3">B6: Time period</td>
  <td>E6: 2023-Q3</td>
  <td>F6: 2023-Q4</td>
  <td>G6: 2024-Q1</td>
  <td>H6: 2024-Q2</td>
  <td>I6: 2024-Q3</td>
  <td>J6:</td>
</tr>
<tr>
  <td>A7:</td>
  <td>B7: Combined transaction</td>
  <td>C7: Combined unit of measure</td>
  <td>D7:</td>
  <td>E7:</td>
  <td>F7:</td>
  <td>G7:</td>
  <td>H7:</td>
  <td>I7:</td>
  <td>J7:</td>
</tr>
<tr>
  <td>A8:</td>
  <td colspan="7">B8: Adjustment: Calendar and seasonally adjusted</td>
</tr>

\end{verbatim}

\begin{itemize}
    \item HTML + Colspan + RowTag
\end{itemize}
\begin{verbatim}
<tr rownum=3>
  <td>Time period</td>
  <td>2023-Q3</td>
  <td>2023-Q4</td>
  <td>2024-Q1</td>
  <td>2024-Q2</td>
  <td>2024-Q3</td>
  <td></td>
</tr>
<tr>
  <td></td>
  <td>Combined transaction</td>
  <td>Combined unit of measure</td>
  <td></td>
  <td></td>
  <td></td>
  <td></td>
  <td></td>
  <td></td>
  <td></td>
</tr>
<tr>
  <td></td>
  <td colspan="7">Adjustment: Calendar and seasonally adjusted</td>
  <td rowspan="4"></td>
</tr>

\end{verbatim}

\end{document}